%% file: main.tex
\documentclass[11pt]{article}

\usepackage[final]{acl}

\usepackage{times}
\usepackage{latexsym}

\usepackage[T1]{fontenc}

\usepackage[utf8]{inputenc}

\usepackage{microtype}

\usepackage{inconsolata}

\usepackage{graphicx}
\usepackage{algorithm}
\usepackage{algpseudocode}
\usepackage{amsmath,amssymb}   
\usepackage{graphicx}           
\usepackage{xcolor}             
\usepackage{booktabs}       
\usepackage{amsmath}
\usepackage{cases}
\usepackage{bm}
\usepackage{amsmath}
\usepackage{multirow}
\usepackage{cuted}
\usepackage{tcolorbox}
\tcbuselibrary{breakable}
\usepackage{caption}

\usepackage{graphicx}      
\usepackage{booktabs}      
\usepackage{multirow}      
\usepackage{subcaption}    
\usepackage[table]{xcolor} 
\usepackage{tcolorbox}     
\usepackage{enumitem}
\usepackage{fancyvrb}
\usepackage{xurl}
\usepackage[dvipsnames]{xcolor}

\definecolor{myhighlight}{HTML}{E5F5E0}

\usepackage{xspace}
\newcommand{\reflexicoder}{ReflexiCoder\xspace} 

\title{ReflexiCoder: Teaching Large Language Models to Self-Reflect on Generated Code and Self-Correct It via Reinforcement Learning}

\author{
Juyong Jiang$^1$\,$^2$\thanks{Work done during a research internship at NAVER.}\quad 
Jiasi Shen$^2$\thanks{Corresponding authors.}\quad
Sunghun Kim$^1$\quad
Kang Min Yoo$^3$ \\
\textbf{Jeonghoon Kim}$^4$\footnotemark[2]\quad
\textbf{Sungju Kim}$^4$\footnotemark[2]\\
$^1$The Hong Kong University of Science and Technology (Guangzhou) \\
$^2$The Hong Kong University of Science and Technology\\
$^3$Amazon AGI \quad $^4$NAVER Cloud\\
\texttt{csjuyongjiang@gmail.com}, \texttt{\{sjs,hunkim\}@cse.ust.hk}\\
\texttt{kangminy@amazon.com}, 
\texttt{\{jeonghoon.samuel,sungju.kim\}@navercorp.com} 
}

\begin{document}
\maketitle

\begin{abstract}
While Large Language Models (LLMs) have revolutionized code generation, standard ``System 1'' approaches that generate solutions in a single forward pass often hit a performance ceiling on complex algorithmic tasks. Existing iterative refinement strategies attempt to bridge this gap at inference time, yet they predominantly rely on external oracles, execution feedback, or computationally expensive prompt-response cycles. In this work, we propose ReflexiCoder, a novel reinforcement learning (RL) framework that internalizes the structured reasoning trajectory, encompassing initial generation, bug and optimization aware reflection, and self-correction, directly into the model's weights. Unlike prior methods, ReflexiCoder shifts the paradigm from \textit{external-dependent refinement} to \textit{an intrinsic, fully autonomous self-reflection and self-correction capabilities} at inference time. We utilize an RL-only training paradigm with granular reward functions to optimize the entire reflection-correction trajectory, teaching the model \textit{how to debug} without reliance on ground-truth feedback or execution engines at inference time. Extensive experiments across seven benchmarks demonstrate that our ReflexiCoder-8B establishes a new state-of-the-art (SOTA) among leading open-source models in the 1.5B to 14B range, achieving 
\textbf{94.51\%} \textbf{(87.20\%)} on HumanEval (Plus), \textbf{81.80\%} \textbf{(78.57\%)} on MBPP (Plus), 
\textbf{35.00\%} on BigCodeBench, \textbf{52.21\%} on LiveCodeBench, and \textbf{37.34\%} on CodeForces in a \textbf{single-attempt} setting,
rivaling or surpassing proprietary models like GPT-5.1. Notably, our framework is significantly more token-efficient than base models, reducing inference-time compute overhead by approximately \textbf{40\%} through disciplined, efficient reasoning and reflection patterns. 
The source code and data are available at \texttt{\url{https://github.com/juyongjiang/ReflexiCoder}}. 
\end{abstract}

\input{sections/1_introduction}
\input{sections/3_methodology}
\input{sections/4_experiments}

\input{sections/5_conclusion}

\section{Limitations}
Our proposed ReflexiCoder improves reliability by allocating multiple reflection and correction cycles. Even with cycle regulation and efficiency bonuses, the method may increase token usage and latency compared to single pass generation. This trade-off can limit applicability in tight latency settings, and the optimal reflection budget may vary by task difficulty in ways that are hard to predict a priori.
The proposed intrinsic debugging primarily targets algorithmic correctness and local code issues within a single file setting. It does not explicitly address repository level development, long horizon refactoring, dependency management, or interactive debugging with evolving specifications. Extending the trajectory formulation to multi-file contexts and richer tool interfaces remains future work.
We instantiate ReflexiCoder on Qwen3 family models and evaluate on common Python centric benchmarks. While scaling trends are promising, it is unclear how well the same trajectory format and reward shaping transfer to other base models, other programming languages, or domains where correctness cannot be captured by unit tests alone.

\bibliography{ref}

\appendix
\input{sections/appendix}

\end{document}

%% file: sections/1_introduction.tex
\section{Introduction}

\begin{figure*}[t]
    \centering
    \includegraphics[width=\linewidth]{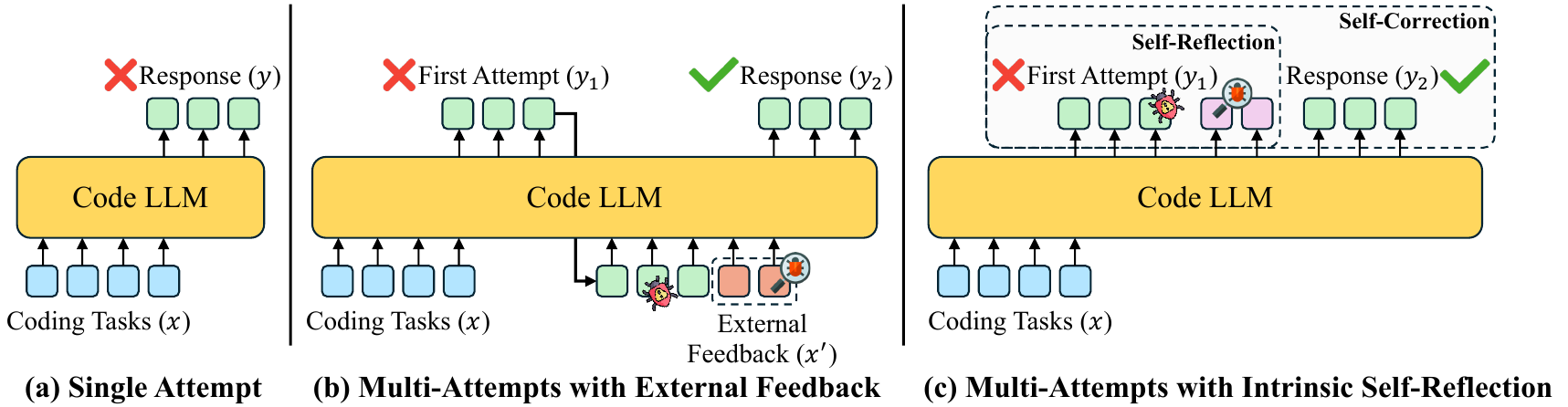}
    \caption{A comparative overview of iterative code refinement workflows at inference time. (a) Existing code LLMs often struggle to generate correct solutions for complex programming tasks on a single attempt. (b) Prior practices mitigate this by relying on external feedback (e.g., compilers, reflection or human oracles). (c) Our proposed \reflexicoder fosters an intrinsic capability to self-reflect and self-correct via a structured reasoning trajectory, eliminating the need for external oracles and environmental interaction.}
\label{fig:reflexicode}
\vspace{-5mm}
\end{figure*}

Large Language Models (LLMs) have revolutionized software engineering, demonstrating exceptional proficiency in translating natural language specifications into executable code \citep{chen2021evaluating,dakhel2023github,jiang2024survey,li2025osvbench}. Despite these advancements, standard ``System 1'' approaches which generate solutions in a single forward pass face an inherent ceiling when tackling complex, multi-step algorithmic problems \citep{li2022competition,chen2023teaching,bairi2023codeplan,wang2024kasa,park2025llamaduo,zhong2024ldb}. In intricate scenarios typical of competitive programming or enterprise-level development, even state-of-the-art models frequently produce plausible-looking but functionally incorrect code on their first attempt.

To mitigate this limitation, recent studies have largely pivoted towards iterative refinement strategies at inference time. These can be broadly categorized into three paradigms: (1) \textbf{Re-ranking}, which samples multiple candidates to select the best one \citep{shi2022natural,li2022competition,chen2022codet,zhang2023coder,ni2023lever}; (2) \textbf{External Repairers}, utilizing separate models to patch errors \citep{gupta2020synthesize,jiang2023impact,zhang2023self}; and (3) \textbf{Feedback-Guided Refinement}, including prompt-based self-reflection such as Reflexion \citep{shinn2024reflexion}, which relies on signals from execution environments or prompting frozen models to iteratively improve code \citep{chen2023teaching,jiang2023selfevolve,zhong2024ldb,madaan2024self}.
While effective, these methods suffer from a critical bottleneck: \textit{dependency on external oracles, environmental interaction, and excessive inference-time token consumption}. In real-world development, comprehensive unit tests are often absent, and the iterative overhead of multiple prompt-response cycles leads to significant latency and computational costs. Furthermore, relying on external signals prevents models from internalizing intrinsic debugging capabilities, the ability to scrutinize and correct one's own logic autonomously.

Inspired by the success of reasoning-intensive models like OpenAI o1 \citep{jaech2024openai,qin2024o1} and DeepSeek-R1 \citep{guo2025deepseek}, which utilize extended inference time to facilitate deeper reasoning, we propose that code generation models should similarly possess an autonomous ``inner monologue'' for debugging. 
We introduce \reflexicoder, a novel Reinforcement Learning (RL) framework designed to internalize the structured reasoning trajectory, encompassing initial reasoning, code generation, reflection for bugs and optimization, and correction, directly into the model's weights.  Unlike prior works \citep{shinn2024reflexion,madaan2024self}, \textbf{\reflexicoder~shifts the paradigm from external-dependent refinement to intrinsic, fully autonomous self-reflection and self-correction capabilities at inference time.} By optimizing the entire correction trajectory itself rather than just the generation policy, \textbf{we teach the model the cognitive skill of ``how to debug'' without reliance on ground-truth feedback or external execution engines at inference time}. 
Figure \ref{fig:reflexicode} compares this paradigm shift with prior practices.

To achieve this, our \reflexicoder~utilizes an RL-only (R1-Zero) training paradigm, bypassing traditional supervised fine-tuning (SFT) to autonomously discover efficient reflection-correction patterns tailored to its own parameter space and problem-solving capabilities 
\citep{guo2025deepseek,li2025system}. We optimize the problem-solving trajectory via granular reward functions that incentivizes both accurate error detection and successful repair. 
It is worth noting that our approach marks a fundamental departure from prior RL methods for code generation, such as CodeRL \citep{le2022coderl}, PPOCoder \citep{shojaee2023execution}, DeepCoder \citep{deepcoder2025}. While existing RL methods strictly optimize the \textit{single-pass generation policy} using execution rewards, \textbf{they fail to cultivate 
the intrinsic reasoning capability to 
identify and analyze potential errors and iteratively correct them autonomously
after an initial attempt.} Our \reflexicoder~uniquely applies RL to optimize the \textit{reflection-correction trajectory itself}, transforming self-debugging from an environment-dependent test loop into an \textit{intrinsic cognitive skill}.

Extensive experiments across seven benchmarks demonstrate the efficacy of our approach. \reflexicoder-8B establishes a new state of the art among leading open-source models, achieving 
\textbf{94.51\%} on HumanEval \citep{chen2021evaluating}, \textbf{35.00\%} on BigCodeBench \citep{zhuo2024bigcodebench}, \textbf{52.21\%} on LiveCodeBench \citep{naman2024livecodebench}, and \textbf{37.34\%} on CodeForces \citep{quan2025codeelo}
in a \textbf{single-attempt} setting. When utilizing our iterative reasoning-reflection setup, referred to as \reflexicoder-8B (Multiple), performance further scales to \textbf{95.73\%}, \textbf{36.84\%}, \textbf{54.12\%}, and \textbf{37.68\%}, respectively, surpassing or remaining competitive with proprietary models like GPT-5.1 \citep{GPT5_OpenAI}.

Notably, we show that these gains do not come from excessive inference-time compute overhead, as
our empirical observations (Section \ref{sec:token_budget}) reveal that \reflexicoder~is \textbf{significantly more token-efficient} than base model, consuming approximately \textbf{40\% fewer tokens} in iterative mode. This is driven by a nearly \textbf{50\% reduction} in reasoning tokens, as our RL training teaches the model to efficiently isolate fundamental logic rather than rambling. Furthermore, \reflexicoder~demonstrates a highly \textit{disciplined reflection} pattern, executing exactly one reflection cycle in virtually all cases.
This ensures that our model achieves superior accuracy at the same computational budget in the single-attempt setting, or at an even lower budget in the multi-attempt setting, effectively transforming self-reflection and self-correction into an efficient, low-latency cognitive process.
In summary, our main contributions are as follows: 
\begin{itemize}
\item We propose \reflexicoder, an RL-based framework that transforms self-reflection and self-correction from an environment-dependent test loop into a fully autonomous, intrinsic model capability, eliminating the need for external feedback at inference time.
\item We formulate the reflection-correction loop as a multi-step trajectory and optimize it via RL. Unlike existing RL methods for code generation, our approach targets the \textit{reflection-correction trajectory}, teaching the model the fundamental logic of self-debugging.
\item Our \reflexicoder-8B significantly outperforms leading open-source models and competes with proprietary models like GPT-5.1. We demonstrate that these gains hold even under fair or reduced token-budget comparisons.
\item We release our source code and data to facilitate future research into the internal self-improvement capabilities of LLMs. 
\end{itemize}

%% file: sections/3_methodology.tex
\begin{figure*}[t]
    \centering
    \includegraphics[width=0.95\linewidth]{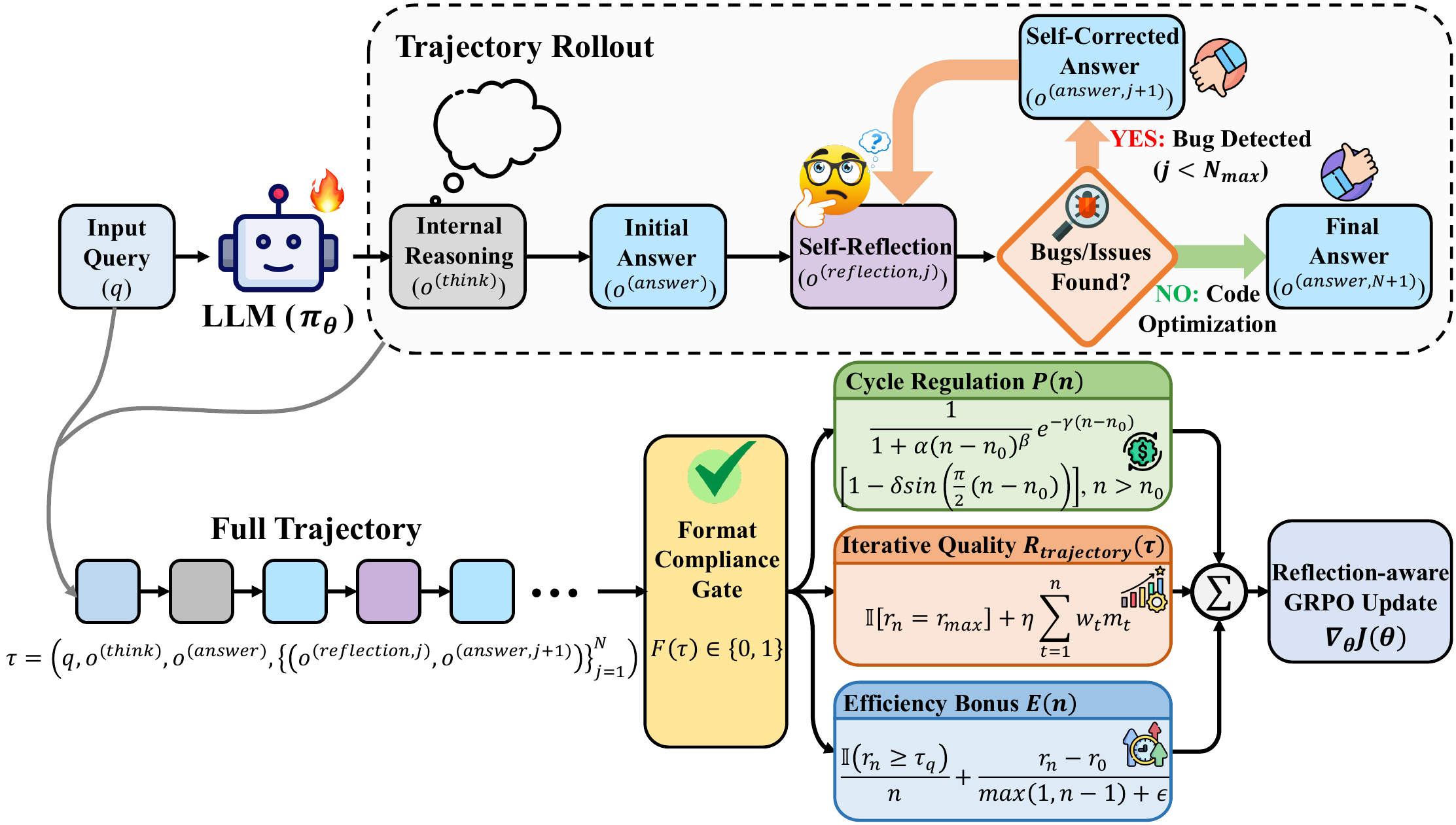}
    \caption{
    The architecture of ReflexiCoder formulates code generation as an RL-optimized intrinsic self-debugging trajectory. 
    A carefully designed composite reward jointly incentivizes reflection quality and correction success.
    }
\label{fig:reflexicode_framework}
\vspace{-5mm}
\end{figure*}
\section{Methodology}\label{sec:method}

In this section, we formalize the proposed ReflexiCoder training pipeline, shown in Figure \ref{fig:reflexicode_framework}, which integrates structured self-reflection and self-correction into LLMs and optimizes the resulting trajectories via RL.

\subsection{Structured Reasoning-Reflection Process}
Let $q \in \mathcal{Q}$ denote a programming-related query, and let an LLM parameterized by $\theta$ produce a sequence of textual outputs in structured segments
\begin{equation}
\begin{aligned}
    o = &(\underbrace{o^{(\text{think})}}_{\text{reasoning}},\ \underbrace{o^{(\text{answer})}}_{\text{answer}},\ \\
    &\underbrace{\{(o^{(\text{reflection},j)}, o^{(\text{answer},j+1)})\}_{j=1}^{n}}_{\text{$n$ reflection cycles}})  
\end{aligned}
\end{equation}
where $n \in \mathbb{N}$ denotes the number of reflection iterations.  
Each reflection-answer pair is constrained to be contiguous and well-formed according to a global format specification $\mathcal{F}$.  

We model the full trajectory corresponding to one prompt-response interaction as
\begin{equation}
\begin{aligned}
    \tau \equiv &\left(q, o^{(\text{think})}, o^{(\text{answer})}, \right. \\ 
    &\left. \{(o^{(\text{reflection},j)}, o^{(\text{answer}, j+1)})\}_{j=1}^{n} \right)  
\end{aligned}
\end{equation}
and define the set of all format-compliant trajectories as $\mathcal{T}_{\mathrm{valid}} = \{ \tau \in \mathcal{T} \mid \Phi(\tau) = \mathcal{F}^{\star} \}$ with $\Phi(\cdot)$ denotes syntax extractor and $\mathcal{F}^{\star}$ the target global format specification, which will be strictly enforced in reward computation.

\subsection{Iterative Reflection Rewards}
\paragraph{Format Compliance Constraints.} 
A fundamental prerequisite for our reinforcement learning setup is that the model's outputs conform exactly to a predetermined structural format. Each generated response must consist of a distinct internal reasoning segment, an initial answer, and a reflection-answer pair for every revision, with the reflection and its subsequent revised answer always appearing together. 
\textit{Additional revision pairs are permitted only when prior reflection identifies new issues, and the total number of reflections must not exceed the specified global limit.}

This structure is not a superficial constraint that our reward mechanisms rely on being able to unambiguously identify each reasoning step, every answer, and the corresponding reflection. Deviations, such as missing segments, incorrect ordering, or unmatched reflection-answer pairs, break the parsing pipeline and undermine the core iterative improvement process. To enforce strict adherence, we introduce a format compliance reward $ F(\tau)$:
\begin{equation}\label{eq:format_reward}
F(\tau) = \mathbb{I}\big[ \tau \in \mathcal{T}_{\mathrm{valid}} \big] \Rightarrow F: \mathcal{T} \to \{0,1\}.
\end{equation}
This binary reward $F(\cdot)$ acts as a gating factor that if $ F(\tau) = 0 $, the total reward for the trajectory is zero, irrespective of content quality. Only trajectories that satisfy the format constraint are eligible for further quality-related reward shaping.

Once format compliance is guaranteed, the reward model incorporates three complementary components, including a smoothly decaying penalty for excessive reflection cycles, a trajectory improvement term that emphasizes progressive quality gains, and an efficiency bonus that rewards significant improvement with minimal iteration.

\paragraph{Cycle Count Regulation.} Reflection cycles inherently present a trade-off between depth and efficiency. Empirically, one to three cycles often yield substantial benefits, such as clarity enhancement, logical coherence, and factual accuracy. Beyond that, gains diminish, and the LLM may waste computational effort or even regress. Let $n \in \mathbb{N} $ denote the total number of reflection cycles, and $n_0$ is the no-penalty depth. When $1 \le n \leq n_0 $, we apply no penalty ($P(n) = 1$), preserving the freedom to engage in ``reasonable depth'' revision. For $ n > n_0 $, rewards ($P(n) \in (0, 1]$) are multiplicatively attenuated by a composite decay term:
\begin{equation}\label{eq:cycle_penalty}
    P(n) =
    \left\{
    \begin{array}{@{}l@{\hspace{-0.5em}}l@{}} 
    1, & \hspace{-2em} 1 \le n \le n_0, \\[4pt]
    \begin{aligned}
    &\frac{1}{1 + \alpha (n - n_0)^{\beta}}
    \cdot e^{-\gamma (n - n_0)} \\
    &\quad \cdot \left[ 1 - \delta \sin\!\left( \frac{\pi}{2}(n - n_0) \right) \right],
    \end{aligned}
    & n > n_0
    \end{array}
    \right.
\end{equation}
where $\alpha > 0$, $\beta > 1$ control polynomial decay strength, $\gamma > 0$ governs exponential attenuation, and $\delta \in (0, 0.3)$ introduces a mild oscillatory perturbation that encourages exploration over nearby iteration depths.  
In multi-turn RL, policies often exhibit trajectory collapse \citep{wang2025ragen,park2026tarot}, for example, repeatedly producing the same incorrect code, or oscillating between two erroneous states. A purely monotonic penalty can exacerbate this issue by encouraging premature termination once the model enters a bad local cycle. By injecting a bounded sinusoidal term, we make the per-step penalty slightly non-stationary across turns, which periodically ``nudges'' the policy away from repetitive local optima and promotes exploration of alternative correction paths.

\paragraph{Iterative Quality Improvement.} Beyond regulating cycle count, the learning objective needs to explicitly encourage sustained improvement in the quality of generated answers. We denote the trajectory of quality scores as $\mathbf{r} = (r_0, r_1, \dots, r_n) \in \mathbb{R}^n$, where $r_t$ represents the quality score of the $t$-th solution obtained through automated execution and validation. Ideally, the optimal trajectory should satisfy $r_0 \le r_1 \le \dots \le r_n$, reflecting a progressive improvement in code quality. 
To emphasize the importance of later improvement stages within a trajectory, we apply exponential time-weighting
\begin{equation}\label{eq:weights}
w_t = \frac{e^{\lambda t}}{\sum_{k=1}^{n} e^{\lambda k}}, \quad \lambda>0
\end{equation}
which yields a normalized vector $\mathbf{w}=(w_1, w_2, \dots, w_n) \in \Delta^n$ over the probability simplex in $\mathbb{R}^n$, with the parameter $\lambda$ controlling the degree to which later iterations are prioritized.
The resulting weights satisfy $w_1 < w_2 < \dots < w_n$, thereby favoring improvements occurring in later stages.

A central challenge in iterative refinement lies in designing a reward signal that captures not only the absolute quality of each answer but also the trajectory's progression. Let $\Delta r_t = r_t - r_{t-1}$ for $t \ge 1$ denote the gains in quality between successive answers. We define the improvement signal $m_t$ using a piecewise formulation:
\begin{equation}\label{eq:improvement_signal}
m_t =
\left\{
\begin{array}{@{}l@{\quad}r@{}}
+f\left(\frac{\Delta r_t}{s}\right) & \Delta r_t > 0, \\[4pt]
+h_{pos} & \hspace{-3em}|\Delta r_t| < \varepsilon \; \text{and} \; |r_{t-1} - r_{\max}| < \varepsilon, \\[4pt]
-g\left(\frac{|\Delta r_t|}{s}\right) & \Delta r_t < 0, \\[4pt]
-h_{neg} & \hspace{-4em}|\Delta r_t| < \varepsilon \; \text{and} \; r_{t-1} < r_{\max}
\end{array}
\right.
\end{equation}
where $s>0$ controls the sensitivity to quality changes, $\varepsilon$ is a small tolerance for numerical stability, $r_{\max}$ denotes the maximum achievable score, and 
$h_{pos}>0$ and $h_{neg}>0$ are constants used to handle stagnation, with $h_{pos}$ providing a bonus when the score has effectively converged near $r_{\max}$ and $h_{neg}$ imposing a penalty when the answer stagnates below $r_{\max}$.
We adopt $\tanh(\cdot)$ for $f(\cdot)$ and $g(\cdot)$ as it provides a smooth mapping from raw score differences to bounded rewards, which facilitates stable policy optimization.

The trajectory-level reward is then defined as
\begin{equation}\label{eq:trajectory_reward}
R_{\mathrm{trajectory}}(\tau) = \underbrace{\mathbb{I}[r_n=r_{max}]}_{\text{final solution}} + \underbrace{\eta\sum_{t=1}^n w_t m_t}_{\text{quality improv.}}, 
\end{equation}
where $r_{max}=1$ means code passes all tests, $\eta > 0$ adjusts the contribution of the improvement signal relative to the absolute quality score.

Notably, this reward design provides positive reinforcement for quality gains, penalizes declines, suppresses stagnation when improvement is still possible, and avoids penalizing the absence of change when the quality is already optimal.
Detailed principles motivating this design are provided in Appendix~\ref{sec:quality_refine}.

\paragraph{Efficiency Reward.} However, solely combining $ P(n) $ and $ R_{\text{trajectory}} $ may lead to undesirable behaviors that the model might overfit to a fixed $n$, ignore task difficulty, or become hypersensitive to noise in $ r_t $. Strong penalties could discourage beneficial exploration, and credit assignment over long horizons remains problematic \citep{parthasarathi2025grpo}. To counter these problems, we introduce an efficiency term:
\begin{equation}\label{eq:efficiency}
E(n) = \underbrace{\frac{\mathbb{I}[r_n \ge \tau_q]}{n}}_{\text{absolute}} + \underbrace{\frac{r_n - r_0}{\max{(1, n - 1)} + \epsilon}}_{\text{relative}}, \; \epsilon > 0
\end{equation}
where $\mathbb{I}[\cdot]$ is the indicator function, $\tau_q$ denotes the required quality threshold, and $\epsilon$ prevents singularities.
This term rewards average quality gain per reflection, encouraging policy to achieve maximal improvement with minimal steps. 

Finally, the overall reward model is:
\begin{equation}\label{eq:overall_reward}
\begin{aligned}
    R_{\text{overall}}(\tau) &= \mathbb{I}[F(\tau)=1] P(n) \big(\varphi R_{\text{trajectory}}(\tau) \\
    & \quad + \psi  E(n) \big) + \xi F(\tau),
\end{aligned}
\end{equation}
where $\varphi$, $\psi$ and $\xi$ control trajectory quality, efficiency bonus, and formatting constraints, respectively.
The reward surface $R_{\mathrm{overall}}$ therefore enforces $\tau \in \mathcal{T}_{\mathrm{valid}}$ and balances \emph{progressive refinement} $R_{\text{trajectory}}$ and \emph{economy of iterations} $E(n)$, formalizing the self-reflection and self-correction objectives in a mathematically explicit manner.
 
In practice, this integrated reward landscape \textbf{allows the learning process to internalize how to reflect effectively across iterations and when to stop}, achieving a disciplined reflection mechanism aligned with the overarching objectives of human. 
The token budget is discussed in Appendix \ref{sec:budget}.

\subsection{Reflection-aware GRPO}\label{sec:main_grpo}
We adopt GRPO objective \cite{guo2025deepseek} for policy $\pi_\theta$ updates, which replaces the value function $V^\pi(s)$ with a group-normalized advantage estimate $\hat{A}(s,a)$, enhancing stability and reducing variance in large action spaces $\mathcal{A}$. 
The detailed formulation is provided in Appendix \ref{sec:grpo}.

%% file: sections/4_experiments.tex
\section{Experiments}
\subsection{Experimental Settings}
\paragraph{Models and Benchmarks}
We instantiate our model by fine-tuning a recent open-source base model, Qwen3-8B \citep{yang2025qwen3}, to obtain our ReflexiCoder-8B.
We evaluate performance across a diverse set of seven widely-used code generation benchmarks, ranging from foundational programming challenges such as HumanEval \citep{chen2021evaluating}, MBPP \citep{austin2021program}, and EvalPlus \citep{liu2023your}, to significantly more complex and competitive programming problems found in BigCodeBench \citep{zhuo2024bigcodebench}, LiveCodeBench \citep{jain2024livecodebench}, and CodeForces \citep{quan2025codeelo}. 

\paragraph{Implementation Details}
We implement our RL pipeline using the TRL~\citep{vonwerra2022trl}.
We train the model for two epochs using our curated open-source dataset of programming problems, derived from \citep{deepcoder2025}. The detailed dataset curation can be found in Appendix \ref{sec:dataset_curation}. 
The no-penalty reflection depth $n_0$ was set to 5. For the cycle count penalty $P(n)$, we used $\alpha=0.1$, $\beta=2.0$, $\gamma=0.05$, and $\delta=0.1$. The exponential weighting for trajectory improvement $R_{\mathrm{trajectory}}(\tau)$ used $\lambda=0.2$, with the improvement signal weight $\eta=0.5$. 
The main reward component weights are $\varphi=0.5$ for trajectory quality $R_{\mathrm{trajectory}}(\tau)$, $\psi=1.0$ for the efficiency bonus $E(n)$, and $\xi=1.0$ for formatting constraints $F(\tau)$. 
All experiments are conducted on a cluster of 8$\times$ NVIDIA H200 GPUs with a per-device batch size of 1. 
More details are provided in Appendix \ref{app:implement}, including the baselines (Section~\ref{sec:baselines}), dataset curation (Section~\ref{sec:dataset_curation}), hyperparameter settings (Section~\ref{sec:hyper_setting}), and the system prompt (Section~\ref{sec:system_prompt}).

\paragraph{Evaluation}
To ensure a rigorous and equitable comparison, we evaluate all baseline models and our ReflexiCoder 
utilizing the EvalChemy \citep{Evalchemy}. 
To decouple the intrinsic quality of the learned policy from gains attributable to increased inference compute, we report results under two configurations with different token budgets.  
First, ReflexiCoder-8B (Single) denotes that the model is evaluated in a single-attempt setting, in which the system prompt (see Appendix Figure \ref{fig:system_prompt}) is removed. 
\textit{This configuration keeps the token budget strictly identical to that of the baseline models, thereby serving as a direct measure of the model's fundamental problem-solving capability.} 
Second, we report ReflexiCoder-8B (Multiple), which utilizes the full iterative reasoning-reflection paradigm enabled by the system prompt. 
Crucially, as discussed in Appendix \ref{sec:budget}, our RL process incentivizes the model to internalize an \textit{optimal trajectory}, encouraging it to generate high-quality, bug-free solutions on the first attempt and require only concise subsequent optimization.

\begin{table*}[t]
\centering
\caption{Main results on seven code generation benchmarks, reporting pass@1 (\%). We compare our ReflexiCoder against leading proprietary and open-source models. Our models establish a new state-of-the-art for open-source models in 1.5B-14B range and demonstrate competitive performance against much larger proprietary models. 
$^*$ denotes results taken from \citep{jiang2024ledex}.   
HE(+) denotes HumanEval(+). BCB is BigCodeBench, LCB is LiveCodeBench, and CF is CodeForces.  
ReflexiCoder-8B (Single) is evaluated in \textit{single-attempt} without the system prompt (see Figure \ref{fig:system_prompt}), while ReflexiCoder-8B (Multiple) uses the full \textit{iterative reasoning-reflection} setup.
We show the score improvement (\textcolor{ForestGreen}{$\pm$}) of our model relative to its base model (Qwen3-8B).
\textbf{Bold} indicates the best performance among open-source and proprietary models, respectively.
}
\label{tab:main_results}
\resizebox{\textwidth}{!}{%
\begin{tabular}{l|l|ccccccc}
\toprule
\textbf{Institution} & \textbf{Model} & \textbf{HE} & \textbf{HE+} & \textbf{MBPP} & \textbf{MBPP+} & \textbf{BCB} & \textbf{LCB} & \textbf{CF} \\
\midrule
\multirow{2}{*}{\includegraphics[height=9pt]{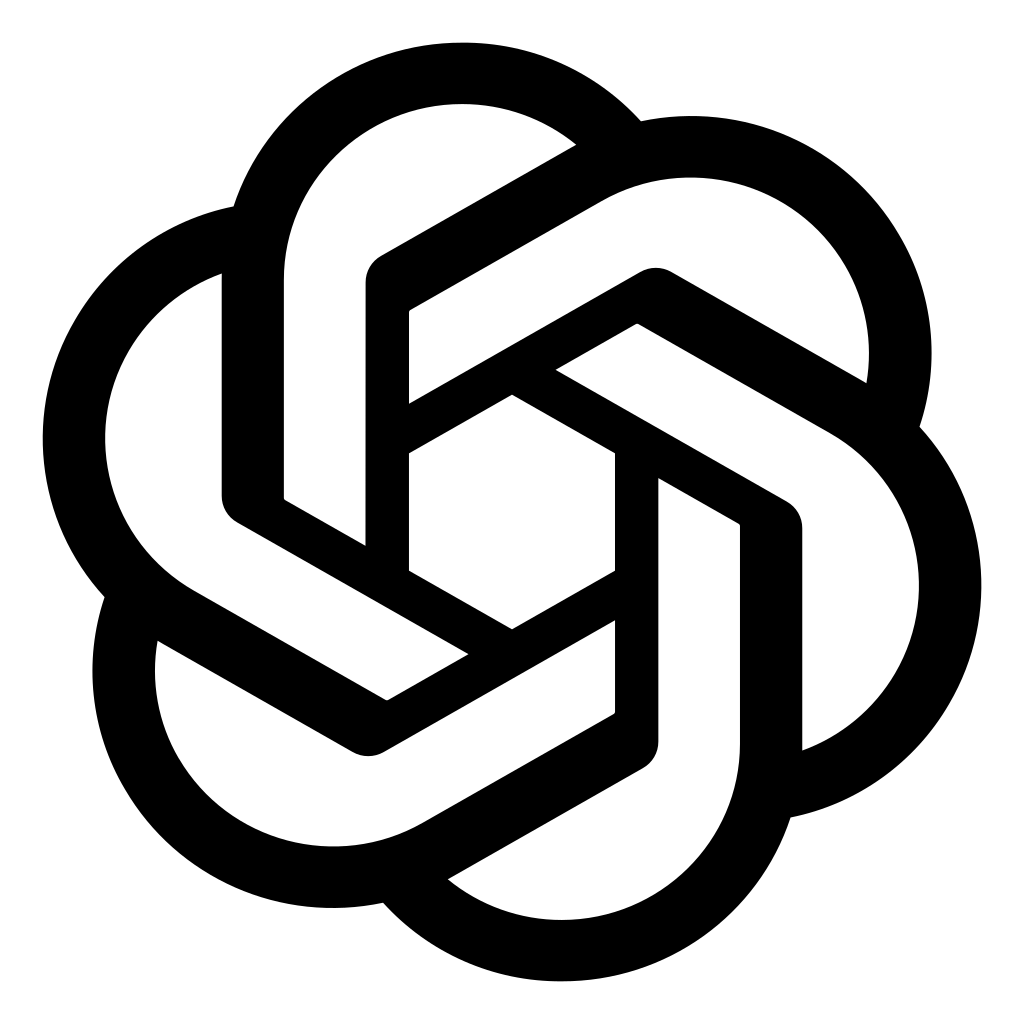} OpenAI} & GPT-5.1 & 95.12 & 87.20 & 84.00 & 79.10 & 39.56 & 48.03 & 34.70 \\
& GPT-4.1 & 96.34 & 78.88 & 85.20 & 79.10 & 41.32 & 42.77 & 31.37 \\
\includegraphics[height=7pt]{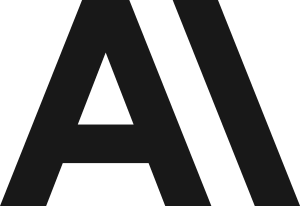} Anthropic & Claude-Sonnet-4.5 & \textbf{98.17} & 77.44 & 76.80 & 75.40 & \textbf{45.00} & 50.78 & \textbf{53.79} \\
\includegraphics[height=9pt]{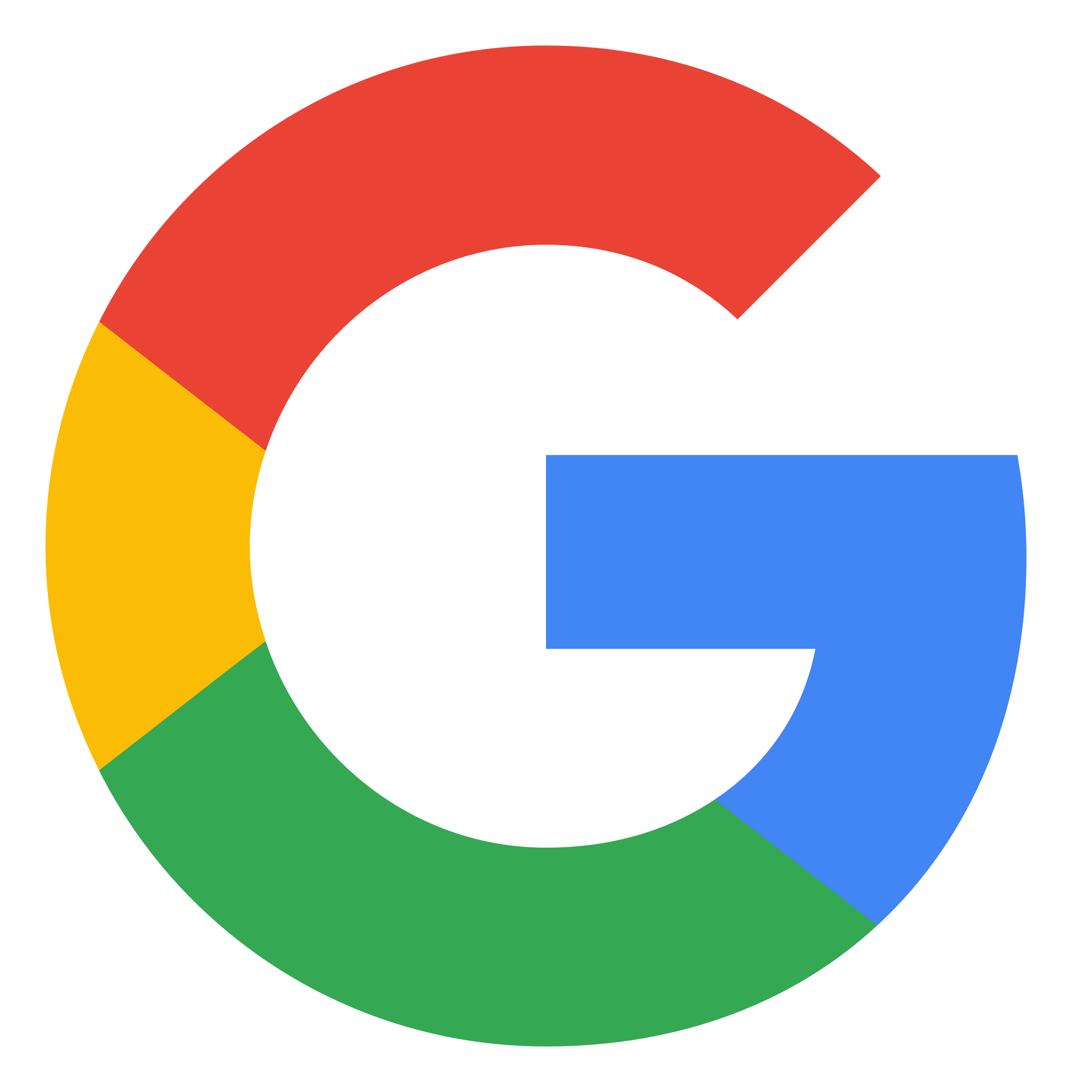} Google & Gemini-2.5-Pro & 97.56 & \textbf{92.07} & \textbf{94.20} & \textbf{84.39} & 41.32 & \textbf{62.01} & 52.40 \\
\midrule
\multirow{2}{*}{\includegraphics[height=12pt]{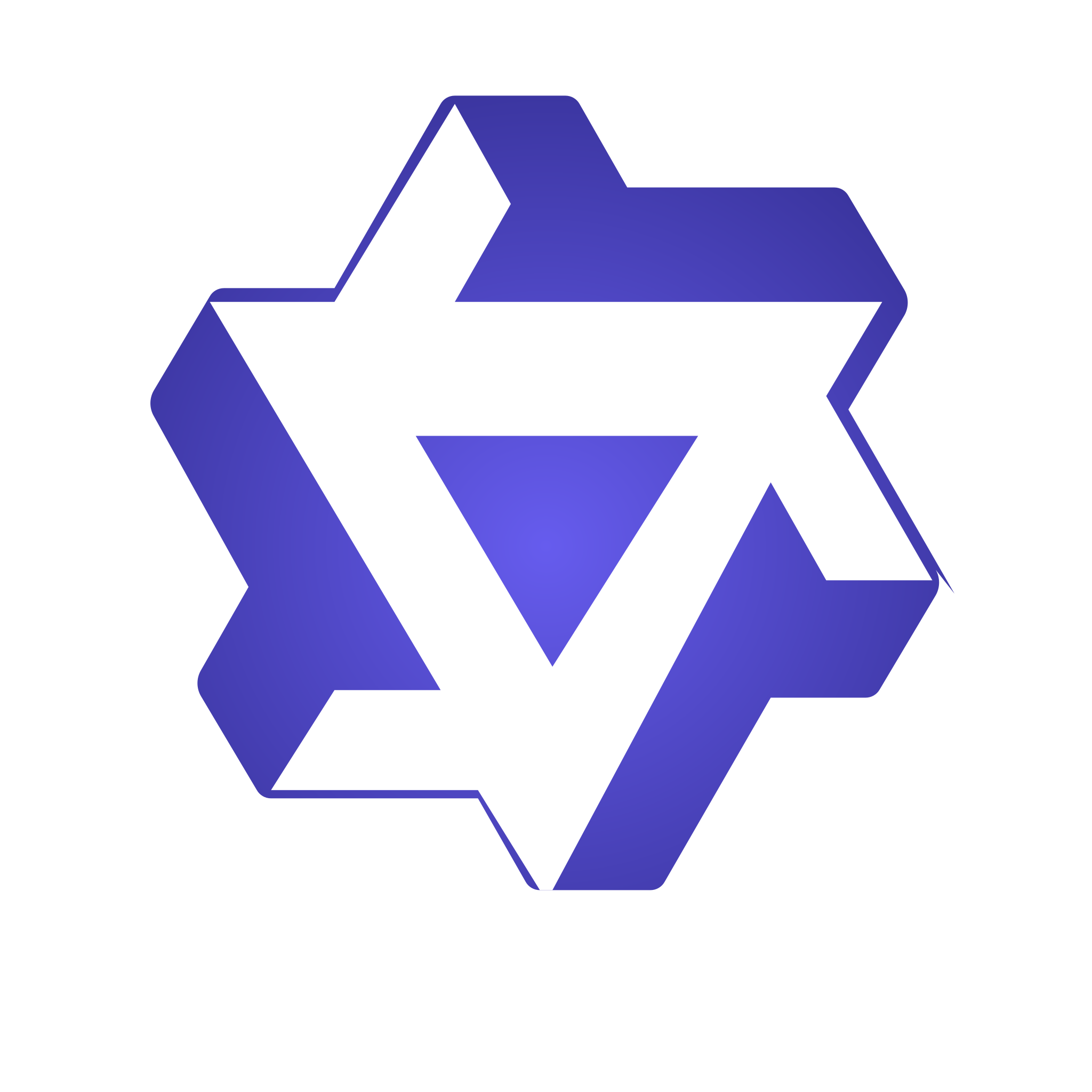} Alibaba} & Qwen3-8B & 89.02 & 80.49 & 70.20 & 70.37 & 32.63 & 37.75 & 23.70 \\
& Qwen2.5-Coder-7B-Instruct & 86.59 & 79.88 & 75.80 & 69.84 & 33.33 & 13.86 & 6.39 \\
\includegraphics[height=9pt]{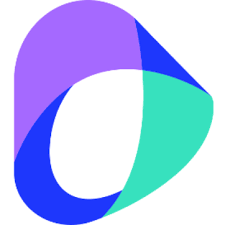} ByteDance & Seed-Coder-8B-Instruct & 85.98 & 80.49 & 68.40 & 72.49 & 36.05 & 21.03 & 4.09  \\
\includegraphics[height=12pt]{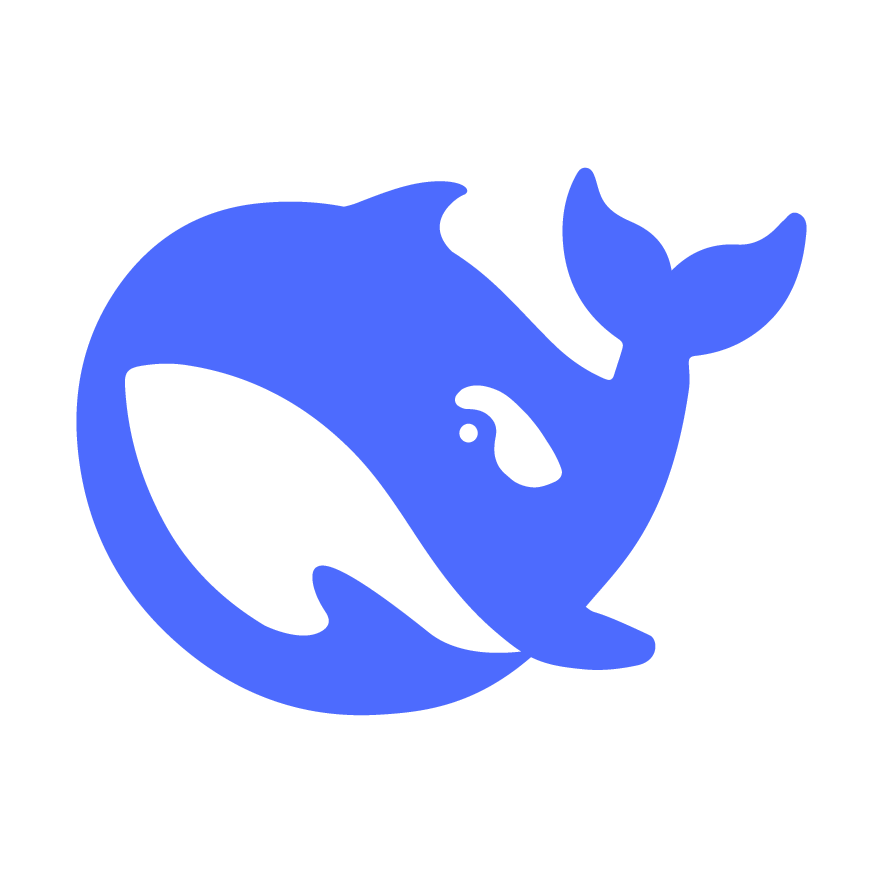} DeepSeek & DeepSeek-Coder-7B-Instruct & 73.17 & 67.07 & 65.00 & 63.76 & 27.02 & 8.24 & 2.56  \\
\includegraphics[height=9pt]{images/logos/google.png} Google & CodeGemma-7B-IT & 54.88 & 41.46 & 53.20 & 54.76 & 25.44 & 8.12 & 1.88 \\
\includegraphics[height=6pt]{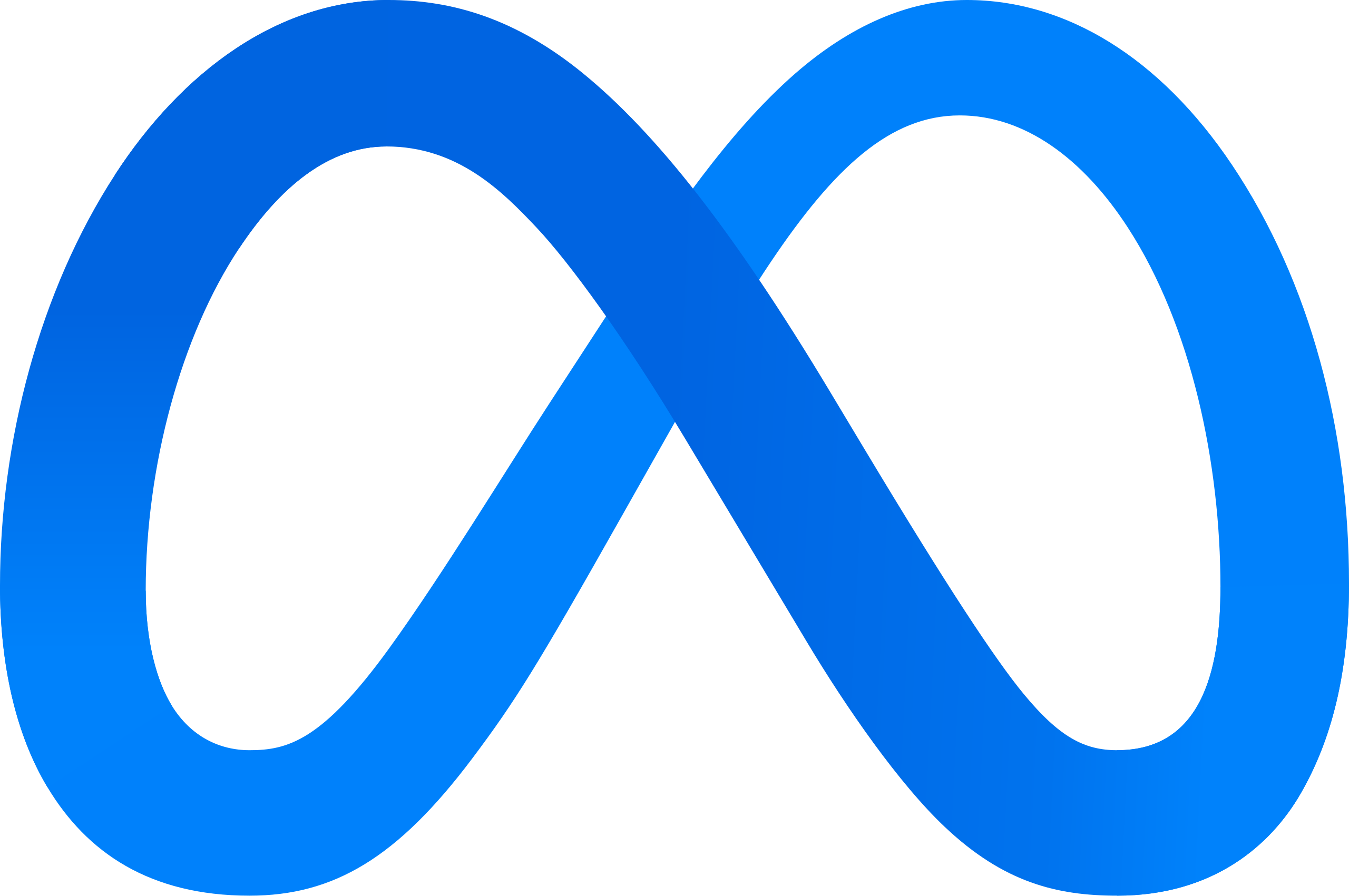} Meta & CodeLlama-7b-Instruct & 30.49 & 26.22 & 38.21$^*$ & 37.18$^*$ & 16.58 & 1.19 & 0.68 \\
\multirow{2}{*}{\includegraphics[height=7pt]{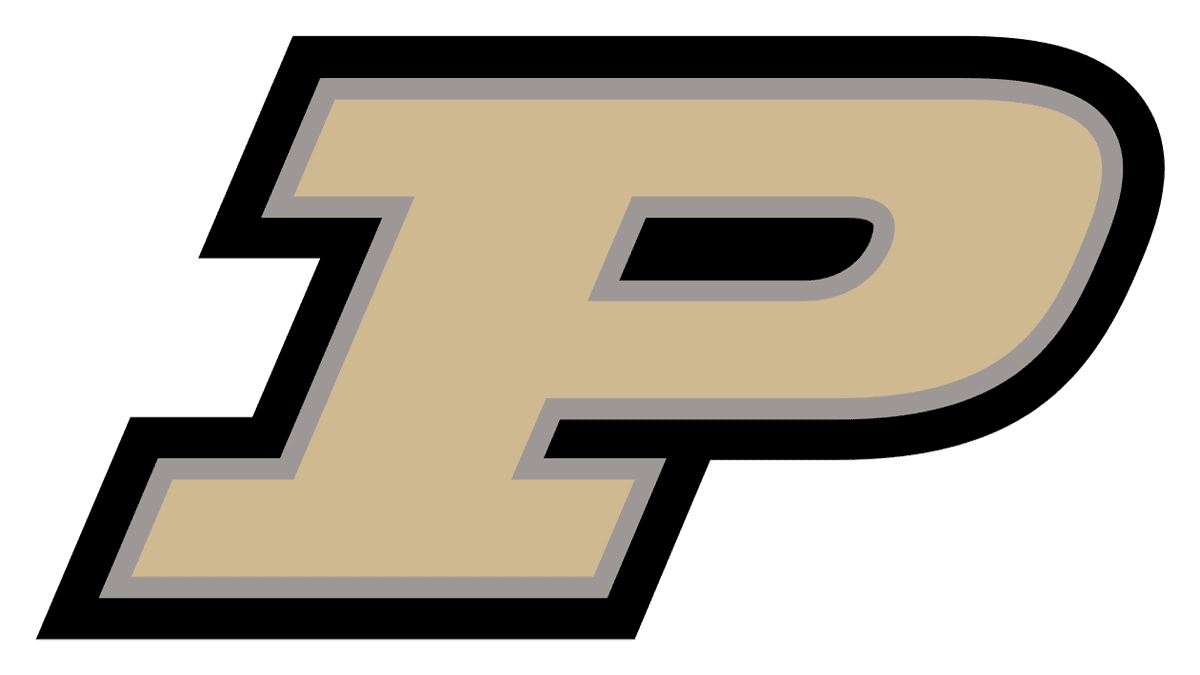} Purdue} 
& LeDex-RL-13B$^*$ & 61.71 & 56.68 & 61.98	& 57.89 & - & - & - \\
& LeDex-RL-7B$^*$ & 55.84 & 50.04 & 57.92 & 52.90 & - & - & - \\
\multirow{2}{*}{\includegraphics[height=9pt]{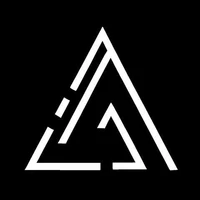} rLLM} 
& DeepCoder-14B-Preview & 80.49 & 75.00 & 63.40 & 67.46 &  28.33 & 34.05 & 14.24 \\
& DeepCoder-1.5B-Preview & 67.07 &	61.59 &	36.00 &	52.12 &	6.84 &	22.94 &	9.89 \\
\midrule
\multirow{2}{*}{\quad Ours} & \textbf{ReflexiCoder-8B (Single)} & \textbf{94.51}{\tiny\textcolor{ForestGreen}{+5.49}} & \textbf{87.20}{\tiny\textcolor{ForestGreen}{+6.71}} & \textbf{81.80}{\tiny\textcolor{ForestGreen}{+11.6}} & \textbf{78.57}{\tiny\textcolor{ForestGreen}{+8.20}} & \textbf{35.00}{\tiny\textcolor{ForestGreen}{+2.37}} & \textbf{52.21}{\tiny\textcolor{ForestGreen}{+14.46}} & \textbf{37.34}{\tiny\textcolor{ForestGreen}{+13.64}}  \\
 & \textbf{ReflexiCoder-8B (Multiple)} & \textbf{95.73}{\tiny\textcolor{ForestGreen}{+6.71}} & \textbf{87.80}{\tiny\textcolor{ForestGreen}{+7.31}} & \textbf{82.00}{\tiny\textcolor{ForestGreen}{+11.8}} & \textbf{79.10}{\tiny\textcolor{ForestGreen}{+8.73}} & \textbf{36.84}{\tiny\textcolor{ForestGreen}{+4.21}} & \textbf{54.12}{\tiny\textcolor{ForestGreen}{+16.37}} & \textbf{37.68}{\tiny\textcolor{ForestGreen}{+13.98}}  \\
\bottomrule
\end{tabular}
}
\end{table*}

\begin{table*}[h!]
\centering
\caption{Ablation study of reward components. Performance is reported as pass@1 (\%) on four benchmarks. 
The results confirm that each component is vital for achieving optimal performance and trajectories.}
\label{tab:ablation}
\resizebox{0.9\textwidth}{!}{
\begin{tabular}{l|cccc}
\toprule
\textbf{Method} & \textbf{HumanEval} & \textbf{BigCodeBench} & \textbf{LiveCodeBench} & \textbf{CodeForces}  \\
\midrule
\textbf{ReflexiCoder-8B (Full)} & \textbf{94.51} & \textbf{35.00} & \textbf{52.21} & \textbf{37.34} \\
\midrule
w/o Format Gating $F(\tau)$ & 84.75 & 32.02 & 39.07 & 24.81  \\
w/o Cycle Regulation $P(n)$ & 92.68 & 33.68 & 52.09 & 35.84  \\
w/o Efficiency Reward $E(n)$ & 91.46 & 33.42 & 42.41 & 29.92  \\
w/o Progressive Imp. $m_t$ & 93.29 & 34.74 & 39.19 & 34.10  \\
\bottomrule
\end{tabular}
}
\end{table*}

\subsection{Main Results}
Table \ref{tab:main_results} presents a comprehensive evaluation of ReflexiCoder-8B across seven diverse code generation benchmarks. Overall, our model establishes a new state-of-the-art for open-source models in the 1.5B to 14B parameter range. Compared to its base model, Qwen3-8B, ReflexiCoder-8B (Single) achieves significant absolute improvements across all metrics, most notably increasing pass@1 accuracy by 14.46\% on LiveCodeBench (LCB) and 13.64\% on CodeForces (CF). These gains on reasoning-intensive tasks underscore the efficacy of our RL paradigm in cultivating deep algorithmic reasoning rather than mere syntax memorization.

Furthermore, ReflexiCoder-8B consistently outperforms both specialized code LLMs, such as Qwen2.5-Coder-7B and Seed-Coder-8B, and recent RL-based models, including LeDex-RL-13B and DeepCoder-14B-Preview. In a strict single-attempt setting, ReflexiCoder-8B (Single) surpasses DeepCoder-14B-Preview by 18.16\% on LCB and 23.10\% on CF, despite having 40\% fewer parameters. This directly validates our core hypothesis that optimizing for the reflection-correction trajectory via RL yields a significantly more robust policy than standard single-pass RL methods. 
Despite its compact 8B scale, ReflexiCoder-8B demonstrates highly competitive performance against much larger proprietary models. Under the iterative setup, ReflexiCoder-8B (Multiple) achieves parity with GPT-5.1 on HumanEval+ (87.80\% vs. 87.20\%) and MBPP+ (79.10\% vs. 79.10\%). Crucially, on the most challenging benchmarks, it outperforms GPT-5.1 by clear margins, scoring 54.12\% (vs. 48.03\%) on LCB and 37.68\% (vs. 34.70\%) on CF. This highlights the effectiveness of our granular reward design in empowering smaller models to handle high-complexity tasks.

Moreover, comparing the Single and Multiple inference settings reveals the intrinsic value of our internalized self-reflection mechanism. By simply activating the system prompt to trigger internal iterations without any external execution feedback, performance scales positively across all benchmarks, such as rising to 95.73\% on HE and 36.84\% on BigCodeBench. 
This confirms that our ReflexiCoder does not merely memorize code patterns but has successfully internalized a generalizable, inference-time self-improvement strategy.

For the remainder of this paper, we report the performance of ReflexiCoder-8B (Single) and refer to it simply as ReflexiCoder-8B unless otherwise specified. 
This choice is strategically motivated: while our \textit{Multiple} variant yields superior results through iterative refinement, we aim to demonstrate that the performance gains of our ReflexiCoder stem from the \textit{intrinsic quality of the foundational policy optimized via RL, rather than simply consuming a higher inference-time token budget}. 
By focusing on the \textit{single-attempt} setting, we eliminate the confounding factor of iterative overhead and prove that our RL paradigm fundamentally enhances the model's reasoning capabilities under a constrained, equivalent computational budget.  

\begin{figure*}[t]
    \centering
    \includegraphics[width=\linewidth]{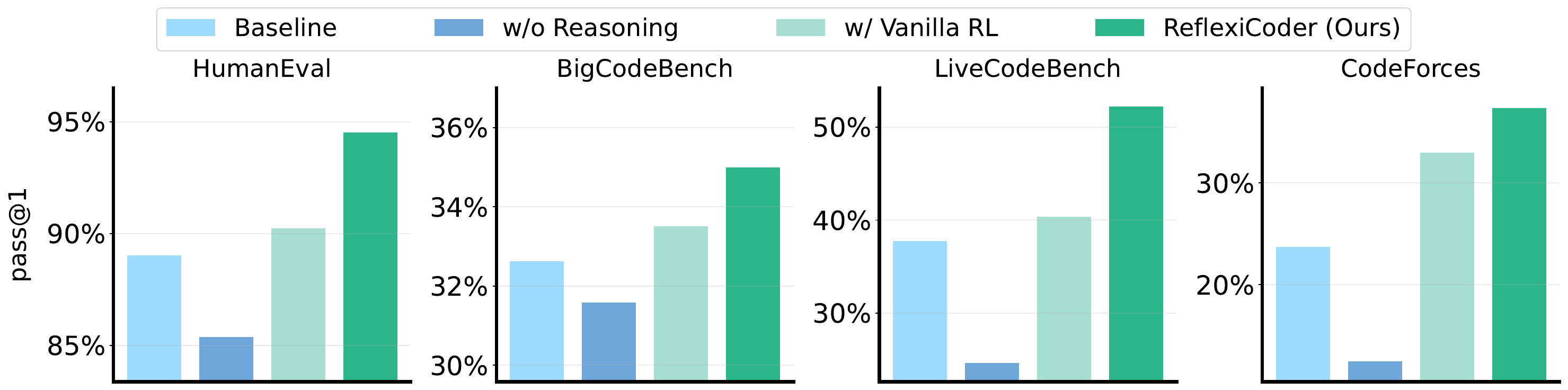}
    \caption{Analysis of the impact of reasoning and reflection. We compare ReflexiCoder against baselines that lack a structured reasoning and reflection step. Performance is pass@1 (\%). The significant performance gap highlights that the structured reasoning-reflection cycle is the key driver of improvement.}
\label{fig:reasoning_reflection}
\vspace{-4mm}
\end{figure*}

\begin{figure*}[t]
    \centering
    \includegraphics[width=\linewidth]{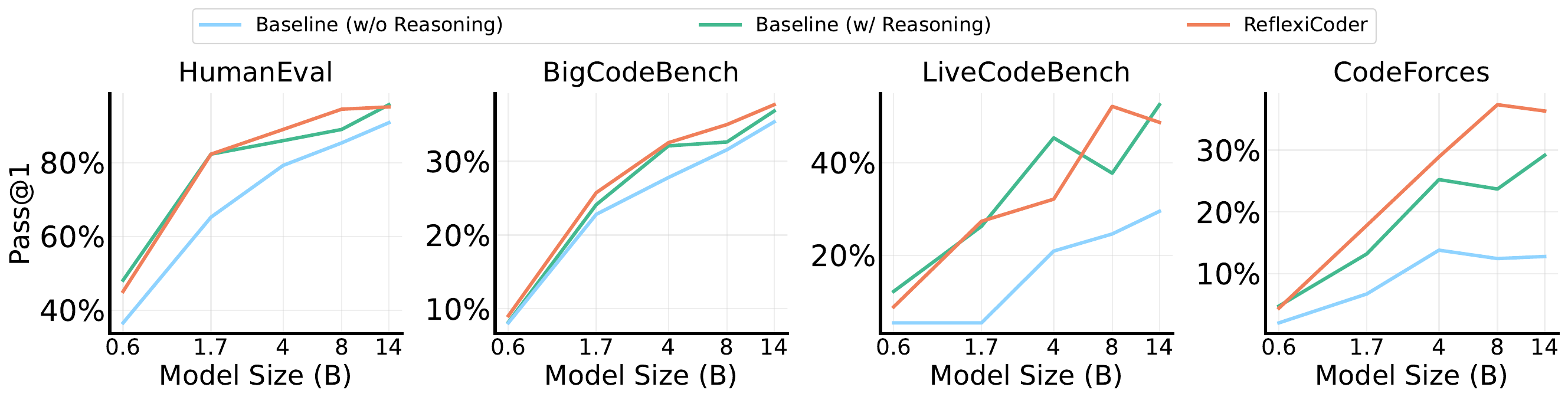}
    \caption{Scaling analysis of ReflexiCoder with model size. Performance is  pass@1 (\%) across four representative benchmarks. The performance grow with model scale, indicating a super-linear benefit.}
\label{fig:model_scaling}
\vspace{-5mm}
\end{figure*}

\subsection{In-depth Analysis and Insights}
In this section, we conduct a series of analyses to understand the emergent behaviors and scalability of ReflexiCoder, with a case study in Appendix \ref{sec:case_study}.

\paragraph{Ablation Study} 
We conduct an ablation study to deconstruct the influence of each component in our composite reward function $R_{\text{overall}}$. 
We train four variants of ReflexiCoder where each variant removes one term from the full reward formulation in Equation \ref{eq:overall_reward}:
(1) Remove Format Gating $F(\tau)$
to quantify the importance of enforcing a strict \textit{reasoning-answer-reflection-answer} format for the learning process.
(2) Remove Cycle Regulation $P(n)$
to test our hypothesis that without regulation, the model may indulge in computationally wasteful or even counter-productive deep reflection, failing to learn when to terminate the process.
(3) Remove Efficiency Reward $E(n)$
to investigate its role in encouraging the model to make more substantial corrections in fewer steps. 
(4) Remove Progressive Improvement $m_t$
from $R_{\mathrm{trajectory}}(\tau)$, relying solely on the absolute quality scores $r_t$, 
to verify that explicitly rewarding the positive delta in quality is crucial for guiding the model towards a monotonically improving trajectory. 
As shown in Table \ref{tab:ablation}, the results demonstrates that each component of our reward designs plays a distinct and indispensable role in achieving the optimal performance and trajectories. 


\paragraph{Impact of Reasoning \& Reflection}
To investigate the source of \reflexicoder's performance gains, we conduct a comparative analysis against three critical baselines: (1) Baseline, the base model without any additional fine-tuning; (2) No Reasoning Pattern, where the base model with non-thinking mode; and (3) Vanilla Outcome-RL, which optimizes the model using a binary pass/fail reward signal without incentivizing the intermediate reflection-correction trajectory. 
The results are summarized in Figure \ref{fig:reasoning_reflection}. 
The substantial performance gains of ReflexiCoder over both the \textit{Non-Reasoning} and \textit{Vanilla RL} baselines with absolute average improvements of 18.64\% and 5.22\%, respectively, across three reasoning-intensive benchmarks demonstrate that the model's success is not merely a byproduct of reasoning and RL, but rather stems from the structured self-reflection and self-correction process. This suggests that our ReflexiCoder has developed an intrinsic debugging capability that mimics human-like cognitive oversight.

\paragraph{Scalability with Model Size} 
We evaluate scalability by training ReflexiCoder on Qwen3 models from 0.6B to 14B. 
Figure \ref{fig:model_scaling} reports average pass@1 over HumanEval, BigCodeBench, LiveCodeBench, and CodeForces for the base model, a \textit{non-reasoning} variant, and our RL-trained ReflexiCoder.
Across all sizes, \textit{non-reasoning} consistently underperforms, with the largest drops on LiveCodeBench and CodeForces, highlighting that intermediate reasoning is critical for algorithmic planning and bug detection. 
In contrast, our ReflexiCoder yields larger gains as model size increases, with especially strong improvements on reasoning-intensive benchmarks. 
This supports our key claim that optimizing the full ``generate, reflect, correct'' trajectory with RL teaches intrinsic self-correction, and larger models can internalize this policy more effectively. Training curves in Appendix Figure \ref{fig:reward_comp} further show faster optimization of $R_{\mathrm{trajectory}}(\tau)$ and $E(n)$ for larger models, indicating more effective and efficient reflection.

\begin{table*}[t]
\centering 
\caption{Token budget statistics and self-reflection count distributions across benchmarks of varying difficulty levels.
The ``Reflection'' column details the frequency distribution of reflection cycles executed per task. The ``Full'' designation refers to the token count of the entire generated response, whereas ``Reasoning'' isolates the specific token footprint of the reasoning process.}
\label{tab:token_budget_stats}
\resizebox{\textwidth}{!}{
\begin{tabular}{lcccccccc}
\toprule
\multirow{2.5}{*}{\textbf{Model}}& \multicolumn{4}{c}{\textbf{HumanEval (164 tasks)}} & \multicolumn{4}{c}{\textbf{BigCodeBench (1,140 tasks)}} \\
\cmidrule(lr){2-5} \cmidrule(lr){6-9}
 & \textbf{Min} & \textbf{Avg} & \textbf{Max} & \textbf{Reflection} & \textbf{Min} & \textbf{Avg} & \textbf{Max} & \textbf{Reflection} \\
\midrule
Qwen2.5-Coder-7B-Instruct  & 61  & 289.18  & 820  & - & 116 & 430.02 & 1104 & -\\
Qwen2.5-Coder-14B-Instruct & 52  & 376.15  & 749  & - & 89  & 329.63 & 770  & -\\
Qwen3-8B (Full)                   & 690 & 4,170.03 & 16,385  & - & 856 & 5,214.01 & 16,385 & - \\
Qwen3-14B (Full)                 & 402 & 3,558.22 & 15,956 & - & 608 & 4,152.02 & 14,572 & - \\
ReflexiCoder (Single) (Full)      & 553 & 3,455.13 & 16,381 & \{0: 164\} & 860 & 3,477.28 & 16,384 & \{0: 1140\}\\
ReflexiCoder (Multiple) (Full)    & 551 & 2,214.51 & 11,107 & \{1: 164\} & 695 & 2,422.33 & 16,385 & \{0: 1, 1: 1139\}\\
\midrule
Qwen3-8B (Reasoning)                   & 0 & 3,133.70 & 15,300  & - & 0 & 4,592.46 & 15,357 & - \\
Qwen3-14B (Reasoning)                 & 0 & 2,744.73 & 13,476 & - & 516 & 3,754.36 & 13,556 & - \\
ReflexiCoder (Single) (Reasoning)      & 0 & 2,701.11 & 13,827 & - & 0 & 2,866.30 & 8,782 & - \\
ReflexiCoder (Multiple) (Reasoning)    & 253 & 1,743.43 & 10,529 & - & 0 & 1,841.67 & 6,919 & - \\
\bottomrule
\end{tabular}
}
\end{table*}

\begin{figure*}[t]
    \centering
    \begin{subfigure}[t]{0.25\linewidth}
        \centering
        \includegraphics[width=\linewidth]{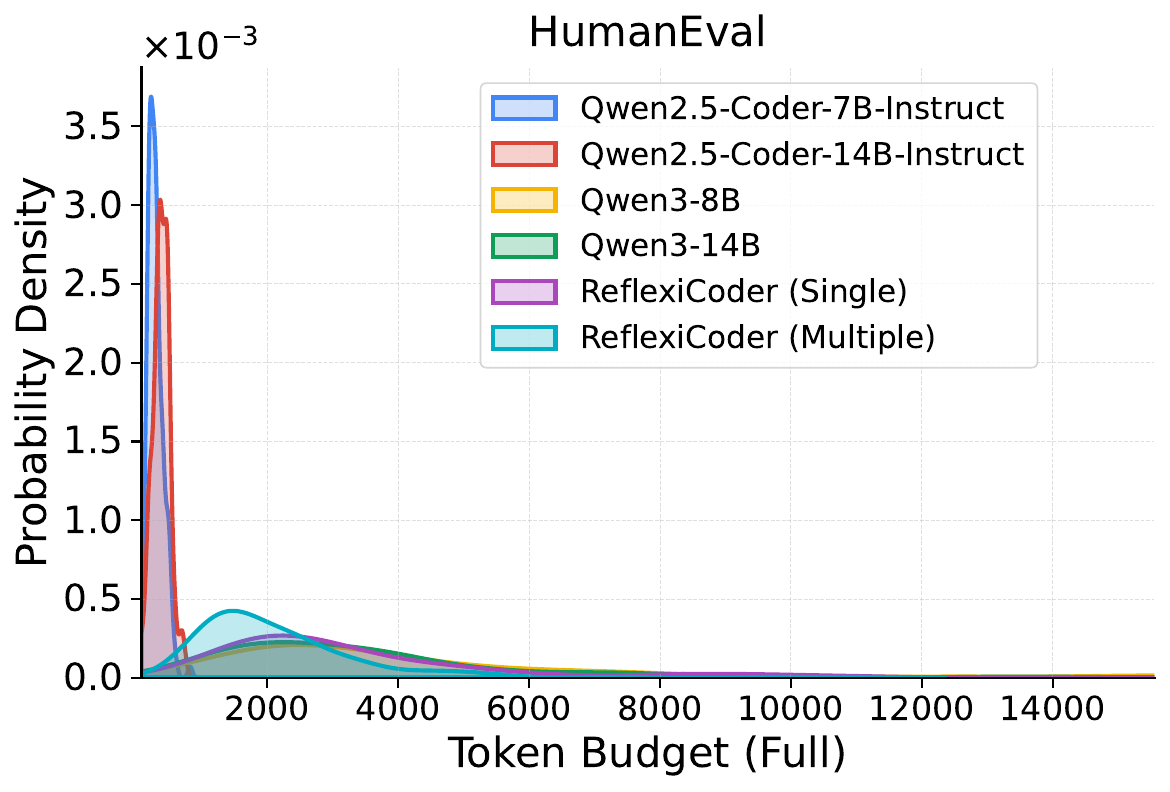}
    \end{subfigure}
    \begin{subfigure}[t]{0.25\linewidth}
        \centering
        \includegraphics[width=\linewidth]{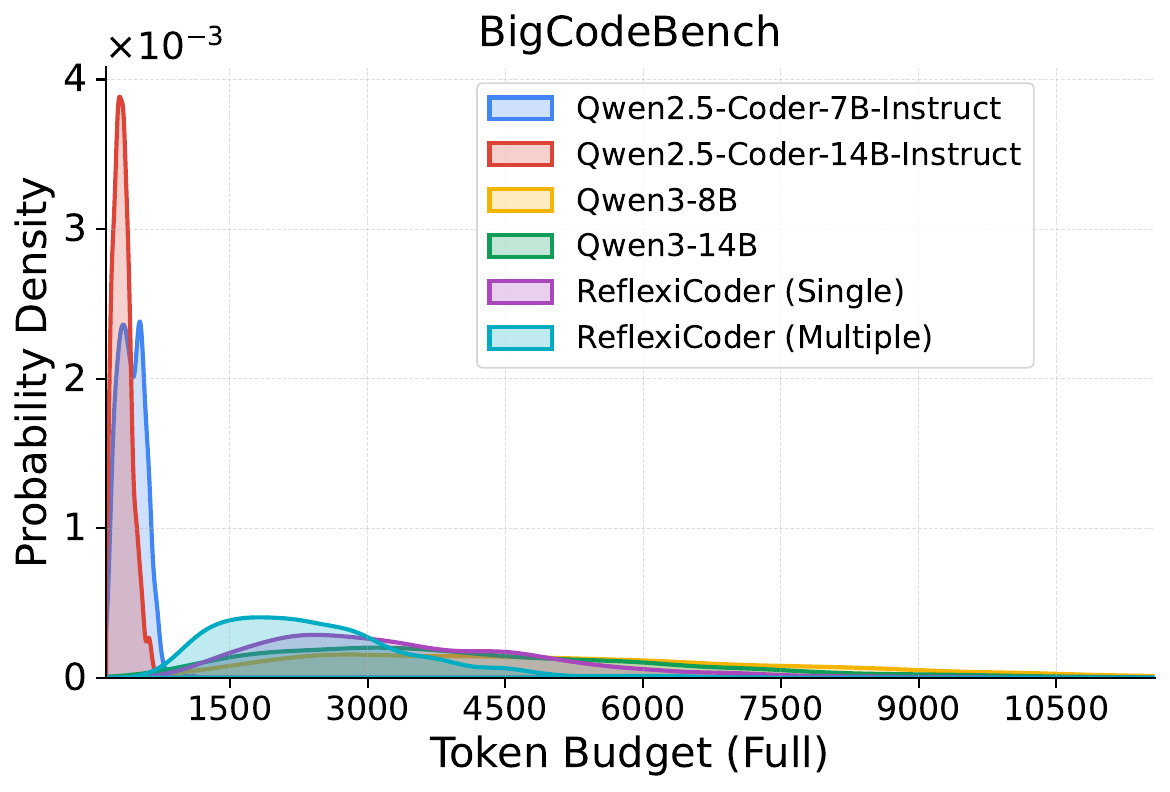}
    \end{subfigure}
    \begin{subfigure}[t]{0.24\linewidth}
        \centering
        \includegraphics[width=\linewidth]{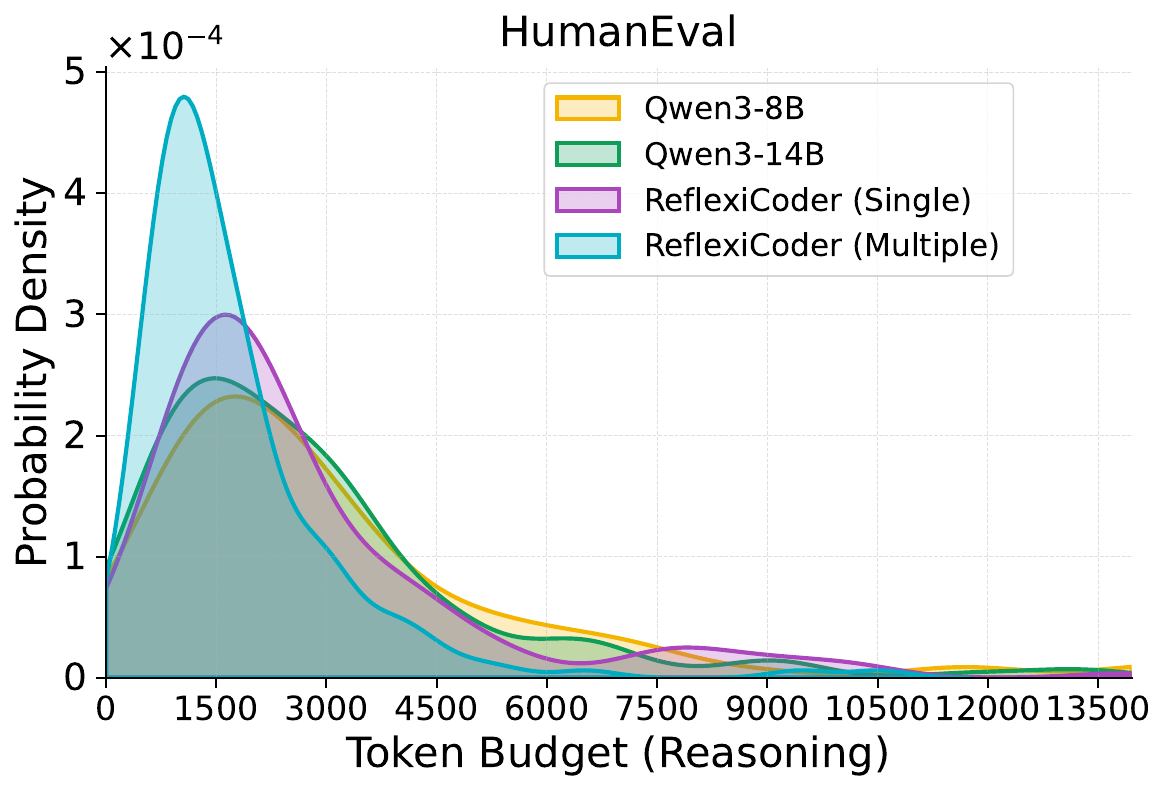}
    \end{subfigure}
    \begin{subfigure}[t]{0.24\linewidth}
        \centering
        \includegraphics[width=\linewidth]{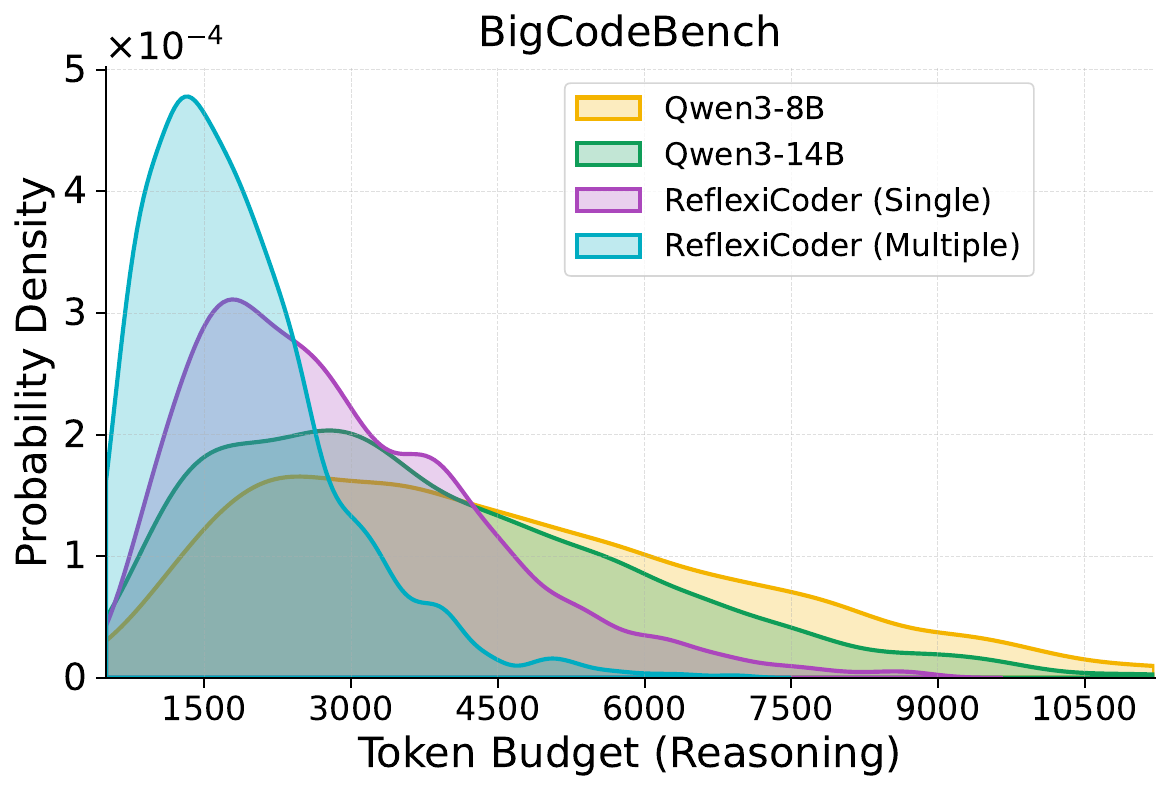}
    \end{subfigure}
    \caption{Probability density distributions of token consumption for both full responses (left) and reasoning segments (right) across the HumanEval and BigCodeBench benchmarks. The reasoning segments correspond to the \texttt{<think>...\null</think>} generation phases. Note that the Qwen2.5-Coder-7B/14B-Instruct baselines are 
    not reasoning models,
    and thus lack corresponding reasoning token budgets.}
    \label{fig:token_budget_dist}
    \vspace{-1.2em}
\end{figure*}

\paragraph{Token Budget Analysis}\label{sec:token_budget} 
To thoroughly investigate the computational overhead of our proposed framework, we analyze the token consumption across different models and settings. As detailed in Table~\ref{tab:token_budget_stats}, both configurations of our model, ReflexiCoder (Single) \textit{without} the system prompt, and ReflexiCoder (Multiple) \textit{with} the system prompt consistently exhibit lower average full token budgets compared to the reasoning baselines, Qwen3-8B and Qwen3-14B. Notably, the ``Reflection'' distribution reveals that ReflexiCoder (Multiple) performs exactly one self-reflection step across nearly all tasks (164/164 on HumanEval and 1,139/1,140 on BigCodeBench). This empirically validates our previous assertion: the model successfully learns the ``optimal trajectory'', synthesizing a correct solution on the first attempt and executing only a single, concise optimization pass. 
Despite engaging in an additional reflection and optimization phase, ReflexiCoder (Multiple) consumes significantly \textit{fewer} total tokens than ReflexiCoder (Single). To uncover the mechanism driving this efficiency, we isolate and analyze the token budget specifically allocated to the reasoning process (denoted as ``Reasoning'' in Table~\ref{tab:token_budget_stats}). A clear hierarchical reduction in reasoning tokens emerges: Qwen3-8B $>$ Qwen3-14B $>$ ReflexiCoder (Single) $>$ ReflexiCoder (Multiple). This demonstrates that our RL training paradigm profoundly enhances the model's core deductive capabilities. By internalizing a more structured and goal-directed reasoning process, ReflexiCoder efficiently isolates the critical logic required to solve the problem, thereby eliminating redundant exploration and verbalization.

These statistical findings are further corroborated by the visual distribution of token budgets shown in Figure~\ref{fig:token_budget_dist}. The probability density curves for both the full response and the reasoning segments illustrate a distinct leftward shift for ReflexiCoder compared to the Qwen3 baselines, indicating a strong concentration toward lower token budgets. Specifically, the distribution for ReflexiCoder (Multiple) exhibits the sharpest peak at the lowest token range across both HumanEval and BigCodeBench. 

Therefore, the dynamic self-reflection and self-correction mechanisms \textbf{do not} impose a computational burden. On the contrary, the highly effective RL optimization suppresses inefficient generation and improves reasoning capabilities, allowing the model to achieve superior performance with a substantially more economical token consumption.

%% file: sections/5_conclusion.tex
\section{Conclusion} 
In this work, we propose ReflexiCoder, a framework that leverages reinforcement learning to teach LLMs intrinsic self-reflection and self-correction without relying on external oracles and environmental interaction at inference time. By formulating debugging as a trainable decision-making trajectory, our ReflexiCoder-8B achieves state-of-the-art performance among open-source models in the 1.5B to 14B range and surpasses or remains competitive with proprietary models like GPT-5.1 on complex reasoning benchmarks. 
These results demonstrate that optimizing internal self-debugging capabilities via RL is a scalable and effective direction for the next generation of reliable code LLMs.

%% file: sections/appendix.tex

\input{sections/2_relatedwork}

\section{Reward Design Principles}\label{sec:quality_refine}
This trajectory-level quality reward is motivated by the intrinsic requirements of iterative refinement tasks in reinforcement learning. The reward structure serves three interrelated objectives as follows:
\begin{itemize}
    \item First, it promotes continuous improvement of answers by providing positive reinforcement for any quality gain and proportional scaling with the magnitude of improvement. 
    \item Second, it penalizes decline in quality and also discourages stagnation when the performance is below the maximum achievable level, thereby maintaining the incentive to search for better solutions. 
    \item Third, it preserves stability once the quality has reached its maximum by avoiding penalties for lack of improvement at that point. 
\end{itemize}
This combination of principles balances the drive for progress with the preservation of optimal states, preventing policies from sacrificing existing high-quality answers in pursuit of transient improvement signals or prematurely halting refinement before reaching optimal performance.

\section{Reflection-aware GRPO}\label{sec:grpo}
For a given prompt $q$ under the old policy $\pi_{\theta_{\text{old}}}$, we sample a group of $G$ outputs $\{ o_i \}_{i=1}^G$, each evaluated with the proposed reward $R_{\text{overall}}(\tau_i)$. The group-relative normalized advantage for the $i$-th trajectory is
\begin{equation}
    \hat{A}_i = \frac{R_{\text{overall}}(\tau_i) - \mu_R}{\sigma_R}
\end{equation}
where $\mu_R = \frac{1}{G} \sum_{j=1}^G R_{\text{overall}}(\tau_j), \sigma_R = \left(\frac{1}{G} \sum_{j=1}^G \left(R_{\text{overall}}(\tau_j) - \mu_R\right)^2\right)^{\frac{1}{2}}$
Policy optimization then follows the clipped surrogate objective:
\begin{equation}
\begin{aligned}
    \mathcal{J}_{\text{GRPO}}(\theta) = \mathbb{E}_{q \sim \mathcal{D},\, \{o_i\} \sim \pi_{\theta_\text{old}}} [ 
    \frac{1}{G} \sum_{i=1}^G \frac{1}{|o_i|} \sum_{t=1}^{|o_i|} \left( \right.\\\left.
    \min( r_{i,t}(\theta) \hat{A}_i,\, \mathrm{clip}\left(r_{i,t}(\theta), 1-\varepsilon, 1+\varepsilon\right)\hat{A}_i) 
    \right.\\\left.
    - \beta_{\text{KL}}\, D_{\mathrm{KL}}(\pi_\theta \parallel \pi_{\mathrm{ref}}) 
    \right) 
    ],     
\end{aligned}
\end{equation}
\begin{equation}
    r_{i,t}(\theta) = \frac{\pi_\theta(o_{i,t} \mid q, o_{i,<t})}{\pi_{\theta_\text{old}}(o_{i,t} \mid q, o_{i,<t})}. \nonumber
\end{equation}
where $r_{i,t}(\theta)$ denotes per-token likelihood ratio. This formulation preserves GRPO's stability advantages while embedding ReflexiCoder's reflection-aware reward into advantage computation, aligning gradient updates with both code correctness and self-reflection efficiency.

\begin{table*}[h]
\centering
\caption{Reward-related hyperparameters. 
Unless otherwise stated, these values are kept fixed across all tasks and benchmarks for reproducibility.}
\begin{tabular}{lcrl}
\toprule
\textbf{Source} & \textbf{Symbol} & \textbf{Value} & \textbf{Meaning} \\
\midrule
\multirow{5}{*}{Eq.~\ref{eq:cycle_penalty}}
& $\alpha$ & 0.1 & polynomial decay strength \\
& $\beta$ & 2.0 & polynomial decay curvature \\
& $\gamma$ & 0.05 & exponential decay strength \\
& $\delta$ & 0.1 & oscillation magnitude \\
& $n_0$ & 5 & reflection depth with no penalty \\
\midrule
\multirow{1}{*}{Eq.~\ref{eq:weights}}
& $\lambda$ & 0.2 & weight concentration on later iterations \\
\midrule
\multirow{5}{*}{Eq.~\ref{eq:improvement_signal}}
& $s$ & 0.1 & sensitivity to quality changes \\
& $\varepsilon$ & 1e-4 & tolerance for numerical stability \\
& $h_{pos}$ & 0.05 & stagnation penalty \\
& $h_{neg}$ & 1.0 & stagnation penalty \\
& $r_{\max}$ & 1.0 & maximum achievable pass rate \\
\midrule
\multirow{1}{*}{Eq.~\ref{eq:trajectory_reward}}
& $\eta$ & 0.5 & weight of improvement term \\
\midrule
\multirow{2}{*}{Eq.~\ref{eq:efficiency}}
& $\tau_q$ & 1.0 & required quality threshold \\
& $\epsilon$ & 1e-6 & prevents division singularity \\
\midrule
\multirow{3}{*}{Eq.~\ref{eq:overall_reward}}
& $\varphi$ & $0.5$ & weight of trajectory quality term \\
& $\psi$ & $1.0$ & weight of efficiency term \\
& $\xi$ & $1.0$ & weight of formatting constraints term \\
\bottomrule
\end{tabular}
\label{tab:hyperparameter}
\vspace{-2mm}
\end{table*}

\begin{figure*}[t]
    \centering
    \includegraphics[width=\linewidth]{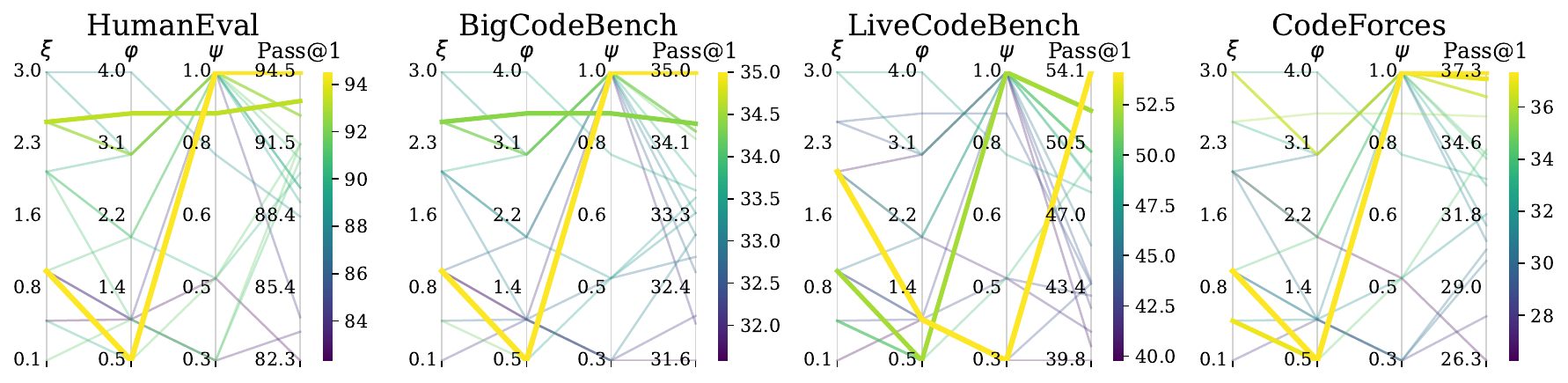}
    \caption{Pass@1 sensitivity to reward weights Format ($\xi$), Iter. Quality ($\varphi$), Efficiency ($\psi$) on four benchmarks.}
\label{fig:hyperparam_sensitivity_weights}
\vspace{-3mm}
\end{figure*}

\section{Implementation Details}\label{app:implement}
In this section, we further illustrate the implementation details as follows. 
\subsection{Baselines}\label{sec:baselines} 
We compare against seven representative code models: Qwen2.5-Coder-7B-Instruct \citep{hui2024qwen2}, Seed-Coder-8B-Instruct \citep{seed2025seed}, DeepSeek-Coder-7B-Instruct \citep{guo2024deepseek}, CodeGemma-7B-IT \citep{team2024codegemma}, and CodeLlama-7B-Instruct \citep{roziere2023code}, LeDex-RL-7B \citep{jiang2024ledex}, and DeepCoder-14B-Preview \citep{deepcoder2025}.   
For the six open-weights models, we perform evaluation using the official chat templates and generation settings recommended by the authors. 
However, as the weights for LeDex-RL-7B are not publicly accessible, we report its results directly from the original paper \citep{jiang2024ledex}.

\begin{figure*}[t]
    \centering
    \includegraphics[width=\linewidth]{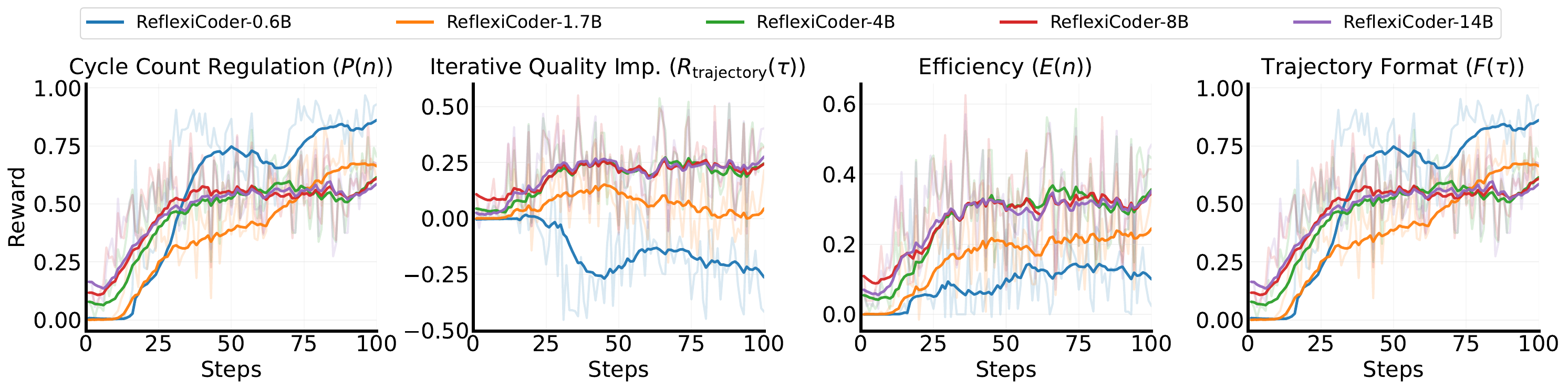}
    \caption{Training dynamics of key reward components and behaviors for different model sizes. Larger models learn to achieve higher progressive improvement and efficiency rewards more quickly. They also converge to a more optimal number of reflection cycles, whereas smaller models struggle to fully optimize the complex reward landscape.}
\label{fig:reward_comp}
\vspace{-5mm}
\end{figure*}

\begin{figure*}[t]
    \centering
    \includegraphics[width=\linewidth]{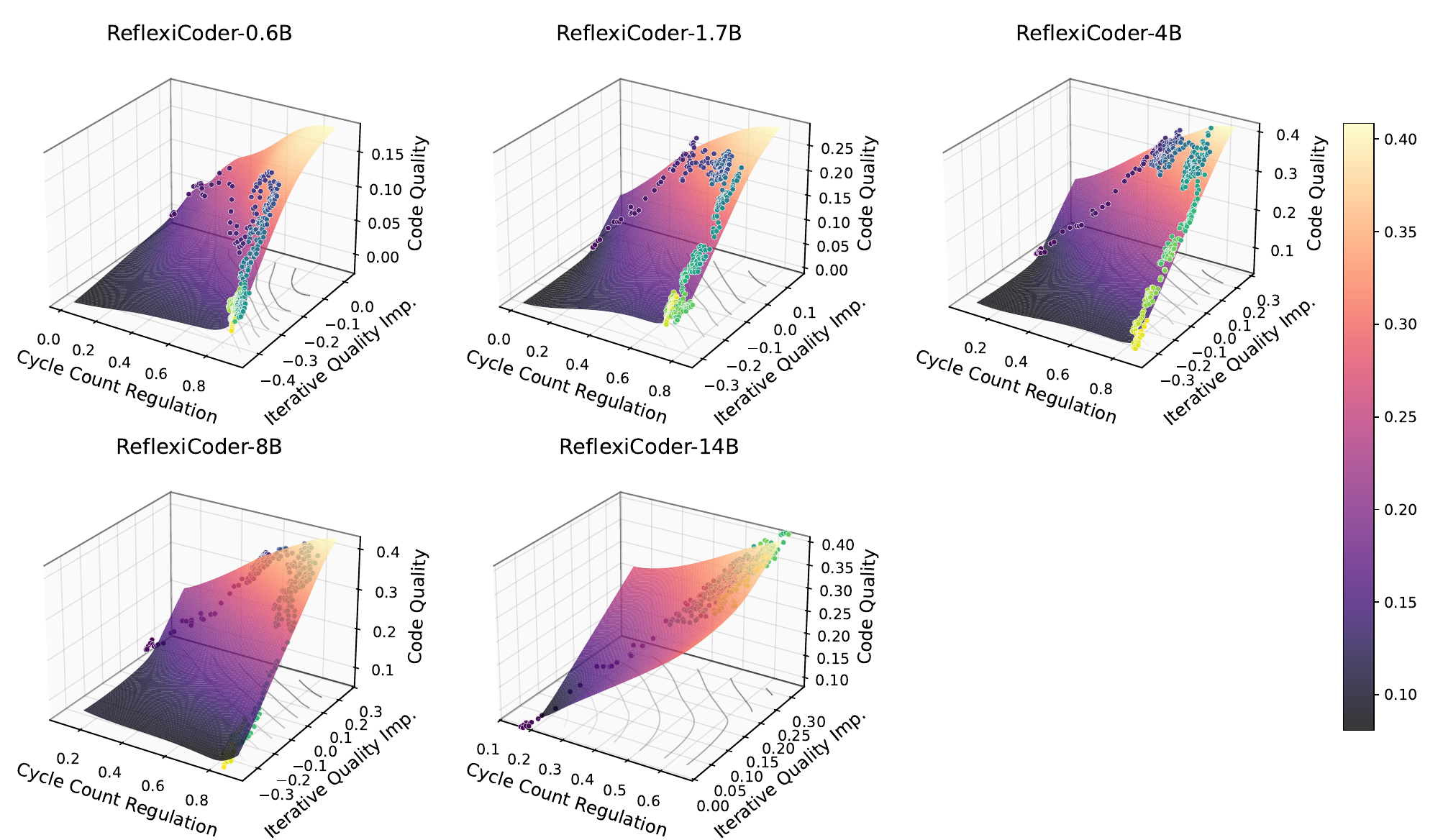}
    \caption{Reward surface induced by shaping terms. 
    For each model scale (0.6B-14B), we fit a smooth 2D surface $ \hat{z}=f(x,y) $ via RBF regression, where $x$ is the cycle-count regulation reward $P(n)$, $y$ is the iterative quality reward $R_{\text{trajectory}}$, and $z$ is final code-quality reward (reward on the last generated code). 
    The surface height and colormap jointly encode the predicted $\hat{z}$ (higher/brighter indicates better code quality), while overlaid points denote the observed training samples $(x_t,y_t,z_t)$ colored by training step; grey contour projections highlight local gradients. Consistent high-$\hat{z}$ regions across scales indicate regimes where shaping terms synergistically improve code quality, whereas steep slopes/contour crowding reveal sensitivity to the corresponding reward component.
    }
\label{fig:reward_correlation}
\vspace{-1mm}
\end{figure*}

\subsection{Dataset Curation}\label{sec:dataset_curation}  
Our training dataset is derived from the open-source DeepCoder training corpus \citep{deepcoder2025}. We directly use four subsets released by DeepCoder: TACO-Verified (7,436 problems), LiveCodeBench (599), CodeForces (6,128), and LeetCode (2,641). 
We follow DeepCoder's released preprocessing, including quality filtering and decontamination, to ensure reliable problem statements and to reduce overlap with standard evaluation benchmarks. 
The LiveCodeBench portion of our training data only contains problems submitted between May 1, 2023 and July 31, 2024. We additionally verify that it does not overlap with the LiveCodeBench v5 test split used by the EvalChemy \footnote{\href{https://github.com/mlfoundations/evalchemy}{https://github.com/mlfoundations/evalchemy}} evaluation framework, whose problems fall in the time window August 1, 2024 to February 1, 2025.

\subsection{Hyperparameters Settings}\label{sec:hyper_setting} 
Our \reflexicoder optimizes a multi-step self-reflection and self-correction trajectory with a composite reward. For clarity and reproducibility, we list all reward-related coefficients and thresholds in Table~\ref{tab:hyperparameter}. Unless otherwise specified, we keep these hyperparameters fixed across all tasks and benchmarks. 
In the sensitivity analysis in Section \ref{sec:sensitivity_analysis}, we observe that final performance is stable over a broad range of values, suggesting that the proposed RL objective does not rely on brittle tuning.

\subsection{System Prompt}\label{sec:system_prompt} 
To mitigate hallucination and logic errors in code generation, we define a structured interaction protocol. As detailed in Figure \ref{fig:system_prompt}, the system prompt instructs the LLM to follow a strict verification loop. Before delivering the final output, the model must: (i) perform a first-pass analysis, (ii) reflect on potential edge cases, and (iii) execute a fix or optimization only if the previous version is confirmed functional. This design ensures that efficiency improvements never sacrifice correctness, a critical safeguard for automated code generation tasks.

\section{Hyperparameter Analysis}\label{sec:sensitivity_analysis}  
We evaluate the sensitivity of \reflexicoder to the reward weights in Eq.~\ref{eq:overall_reward}: trajectory quality $\varphi$, efficiency bonus $\psi$, and format constraint $\xi$. Figure~\ref{fig:hyperparam_sensitivity_weights} shows that performance is robust across a wide range of settings, indicating a stable RL objective. 
The best overall configuration is $\langle\xi,\varphi,\psi\rangle=\langle1.0,0.5,1.0\rangle$, which yields the strongest and most consistent results across all benchmarks, with particularly large gains on reasoning-intensive tasks (LiveCodeBench and CodeForces). Notably, overly large $\varphi$ degrades performance, suggesting that the improvement comes from reward-aligned, effective self-reflection that enables repair, rather than longer or more elaborate reflection. Increasing $\psi$ generally helps, supporting the need to explicitly encourage efficient multi-step self-correction.

\section{Training Dynamics and Reward Scaling}

In this section, we provide a detailed analysis of the reward-learning dynamics across different model scales, as referenced in the main text. 
To investigate how model capacity influences the optimization of the reflection policy, we track the progression of the key reward components during the RL fine-tuning process.

As illustrated in Figure \ref{fig:reward_comp}, we observe that different reward components exhibit distinct scaling behaviors, rather than a uniform ``larger-is-always-better'' trend:
\begin{itemize}
    \item \textbf{Cycle Count Regulation ($P(n)$)}: Larger backbones improve cycle-count control more substantially. In particular, the 0.6B model continues to climb and reaches the highest $P(n)$ by the end of training, while 4B/8B/14B rise quickly but plateau earlier at a lower level. This suggests that smaller models may rely more on increasing/refining the number of cycles to gain reward, whereas larger models learn an adequate stopping behavior earlier.
    \item \textbf{Progressive Improvement ($R_{\text{trajectory}}(\tau)$)}: Scaling is crucial for learning genuine iterative quality gains. The 0.6B model's $R_{\text{trajectory}}(\tau)$ degrades after early training and becomes persistently negative, indicating unstable or even harmful reflection updates. In contrast, models $\ge$4B steadily reach and maintain positive $R_{\text{trajectory}}(\tau)$ (with 8B/14B slightly higher and more stable), suggesting that sufficient capacity is needed to internalize the ``reflect-and-correct'' mechanism without drifting.
    \item \textbf{Efficiency Reward ($E(n)$)}: Larger models achieve higher and more stable efficiency rewards. The 0.6B model remains low throughout training, 1.7B improves but saturates at a moderate value, while 4B/8B/14B converge to a clearly higher plateau. This indicates that scaling helps produce concise yet effective deliberation, reducing redundant or oscillatory reflection traces.
    \item \textbf{Trajectory Format ($F(\tau)$)}: Formatting compliance improves with scale, but the most pronounced gain appears in the smallest model: 0.6B eventually attains the highest $F(\tau)$, while larger models improve quickly and then saturate. This suggests that format adherence is comparatively easy to learn across scales, and may not be the main bottleneck once a model reaches moderate capacity.
\end{itemize} 
Overall, these dynamics reinforce that the gains of our ReflexiCoder are driven by learning structured self-correction policies, especially the ability to generate positive iterative improvements ($R_{\text{trajectory}}(\tau)$) and efficient deliberation ($E(n)$) which emerge reliably only at sufficient model scale, rather than being a mere artifact of longer rollouts or increased sampling.

\section{Interpretation of Reward Shaping}
To understand how our shaping terms drive the performance gains of ReflexiCoder, we visualize the learned reward landscape over training. 
For each model scale, we fit a smooth surface $\hat{z}=f(x,y)$ with RBF regression, where $x=P(n)$ is the cycle-count regulation reward, $y=R_{\text{trajectory}}$ is the iterative quality improvement reward, and $z$ is the final code-quality reward (reward on the last rewritten code). Figure \ref{fig:reward_correlation} reveals three consistent patterns across scales.
The surface height and colormap jointly encode the predicted $\hat{z}$ (higher/brighter indicates better code quality), while overlaid points denote the observed training samples $(x_t,y_t,z_t)$ colored by training step; grey contour projections highlight local gradients. Consistent high-$\hat{z}$ regions across scales indicate regimes where shaping terms synergistically improve code quality, whereas steep slopes/contour crowding reveal sensitivity to the corresponding reward component.
 
Across 0.6B to 14B, the highest $\hat{z}$ concentrates in regions where both $P(n)$ and $R_{\text{trajectory}}$ are strong. This indicates that quality gains are not explained by ``more iterations'' or ``better rewriting'' alone. Instead, the model benefits most when it learns a structured self-debugging trajectory: allocating an appropriate number of reflection cycles while making each revision measurably improve the solution. This supports our framing of self-correction as a multi-step decision process optimized end-to-end by RL.
 
Moreover, the surfaces show that pushing $R_{\text{trajectory}}$ without sufficient $P(n)$ does not reliably yield high $z$. In practice, unconstrained reflection can lead to over-editing, oscillations, or verbose but non-functional changes. The positive gradient along $P(n)$ suggests that cycle-count regulation acts as an implicit budgeting and credit assignment aid, steering the policy toward reflection depths that are most likely to convert into correct final code rather than extended but low-yield ``thinking''.
 
Furthermore, as scale increases, the high-$\hat{z}$ region becomes broader and the attainable $\hat{z}$ increases, indicating that larger models can convert trajectory-level improvement signals into final correctness more consistently. Importantly, the same qualitative landscape persists across scales, suggesting the reward design is not brittle or size-specific. 
This helps explain why our ReflexiCoder's RL training yields robust gains on complex benchmarks: the model is not merely optimizing for a better first-pass solution, but learning an internalizable debugging strategy that generalizes with capacity.

Overall, the visualization provides mechanistic evidence that the performance jump comes from our RL-optimized intrinsic self-reflection and self-correction loop. The shaping terms jointly encourage (i) when to stop reflecting and (ii) how to make corrections that monotonically improve code, reducing reliance on external execution feedback while improving final functional correctness.

\input{images/case_study/case_main}

\section{Token Budget Discussion}\label{sec:budget}
A potential concern regarding our iterative refinement frameworks is the increased computational cost, as multi-round reflection and correction inevitably consume a larger token budget compared to standard single-attempt inference. 
To ensure a fair comparison with baseline models and to demonstrate the intrinsic strength of our ReflexiCoder, we clarify the relationship between our reinforcement learning paradigm and inference-time behavior.

\paragraph{Policy Conditioning via System Prompts.}
The structured reasoning-reflection behavior of ReflexiCoder is conditioned on the \textit{specific} system prompt utilized during RL training. Our training objective encourages the model to internalize the ``Reasoning $\rightarrow$ Answer $\rightarrow$ Reflection $\rightarrow$ Correction'' loop as a specialized operating mode. Crucially, this behavior is not hard-coded but is a learned response to the prompt's instructions. 
\textit{In the absence of this system prompt, ReflexiCoder reverts to the standard inference behavior of its vanilla base model (i.e., Qwen3-8B), producing a single-pass solution without internal reflection and iterative cycles.}

\paragraph{Fair Comparison under Identical Budgets.}
To eliminate any ``unfair'' advantage provided by extra tokens, we evaluate ReflexiCoder on standard benchmarks using a \textit{single-attempt setting} without the iterative system prompt. This ensures that our model operates under the exact same token budget as all baseline models. As demonstrated in Table~\ref{tab:main_results}, our ReflexiCoder consistently outperforms baselines even in this restricted zero-reflection mode. This empirical evidence validates that the proposed RL training pipeline enhances the model's fundamental problem-solving capability rather than simply relying on repeated trials.

\paragraph{Optimal Trajectory Internalization.}\label{sec:optimal_trajectory}
The superior zero-reflection performance is a direct consequence of our reward design (see Eq.~\ref{eq:overall_reward}). \textit{Within the RL environment, the optimal rewards are naturally assigned to trajectories where the initial solution  is correct and requires only a single, brief optimization step.} By optimizing for the maximum expected reward, the model learns to prioritize the ``optimal trajectory'' generating a high-quality, bug-free solution on the first try.

Furthermore, since our efficiency reward (Eq.~\ref{eq:efficiency}) penalizes redundant iterations, the model learns that subsequent reflections should ideally focus on non-functional improvements (e.g., readability or style) rather than fixing logic errors. Consequently, the first-pass success rate (Pass@1) is significantly bolstered, ensuring that the model remains highly competitive and efficient even when computational resources are strictly constrained.

\section{Case Study}\label{sec:case_study}
To provide qualitative insight into the self-correction process, we conduct a case study on a challenging problem from the TACO benchmark. We sample a trajectory generated by ReflexiCoder-8B (Multiple) and manually annotate the errors in the initial solution and the corrections made in subsequent reflection cycles. As illustrated in Figure \ref{fig:case_study}, in Cycle~0, the model produces an initial brute-force implementation to count valid subarrays under the $2^j$ scaling constraint. In Cycle~1, self-reflection detects a correctness bug: the check mistakenly allows equal consecutive scaled values (non-decreasing), and is corrected to enforce strict increase by changing \texttt{<} to \texttt{<=}. In Cycle~2, the model performs an optimization-only revision by precomputing powers of two and reducing redundant computations, improving efficiency and readability while preserving correctness. 

\input{sections/system_prompt}

%% file: sections/2_relatedwork.tex
\section{Related Work}\label{sec:related_work} 
\paragraph{Iterative Refinement with External Feedback.}
Recent advancements suggest that code generation is fundamentally an iterative process rather than a single-turn translation task \citep{shinn2024reflexion,zhuo2025bigcodearena}.  
A prevalent paradigm prompts frozen models to iteratively refine their outputs using external feedback. For example, Self-Debugging \citep{chen2023teaching} and LDB \citep{zhong2024ldb} revise code based on execution traces or unit test results, while Self-Evolve \citep{jiang2023selfevolve} and Reflexion \citep{shinn2024reflexion} incorporate critic-style feedback from external evaluators or self-generated reflections.
LATS \citep{zhou2023language} further augments this paradigm by coupling refinement with Monte Carlo Tree Search (MCTS) guided by external value estimates. 
\textbf{Despite their promise, these methods depend on high-quality external oracles (e.g., compilers, test suites, or critic models), which may be unavailable or expensive in real-world deployment.} 
Unlike these methods, our \reflexicoder internalizes self-debugging by learning to self-correct from intrinsic self-reflection signals, eliminating the need for external oracles (e.g., execution environments or separate critic models) at inference time.

\paragraph{RL for Code and Reasoning.} 
Reinforcement Learning (RL) has been widely adopted to align LLMs with functional correctness. CodeRL \citep{le2022coderl} and PPOCoder \citep{shojaee2023execution} leverage actor-critic architectures to optimize models using compiler feedback or unit test pass rates as reward signals. DeepCoder \citep{deepcoder2025} further explores multi-stage RL to enhance code generation. 
However, these methods typically optimize the \textit{single-pass generation policy} with execution rewards, but \textbf{fail to cultivate the intrinsic reasoning capability to identify and analyze potential errors and to iteratively correct them autonomously after an initial attempt.} 
Recently, reasoning-oriented models like 
DeepSeek-R1 \citep{guo2025deepseek} have demonstrated that extending inference time with intrinsic chain-of-thought (CoT) can significantly boost performance. 
While these models illustrate the potential of test-time scaling, supervising or incentivizing the specific structure of ``self-reflection and self-correction'' for code remains underexplored. \reflexicoder addresses this gap by formulating the self-debugging loop as a structured trajectory and optimizing it via RL, allowing the model to autonomously discover effective strategies for error localization and correction.

%% file: images/case_study/case_main.tex
\begin{figure*}[t]
\centering
\begin{tcolorbox}[
    title=Case Study: Self-Reflection and Correction on a TACO Problem,
]
\textbf{Problem:} Given an array $a$ of length $n$ and an integer $k$, find the number of indices $1 \leq i \leq n - k$ such that the subarray $[a_i, \dots, a_{i+k}]$ with length $k+1$ (not with length $k$) has the following property: ... 
More formally, count the number of indices $1 \leq i \leq n - k$ such that $2^0 \cdot a_i < 2^1 \cdot a_{i+1} < 2^2 \cdot a_{i+2} < \dots < 2^k \cdot a_{i+k}.$.
\hrule\vspace{2mm}
\textbf{Cycle 0: Reasoning and Generation}
\begin{verbatim}
<think>
Okay, I need to solve this programming problem. Let's read the question carefully...
</think>
<answer>
...
                    if val < (a[i + j - 1] * (1 << (j - 1))):
...
</answer>
\end{verbatim}
\vspace{-1mm}
\textbf{Cycle 1: Reflection and Correction}
\vspace{-2mm}
\begin{Verbatim}[commandchars=\\\{\}]
<reflection>\textcolor{Red}{STATUS: BUG\_DETECTED}
The logic inside the inner loop currently uses the condition if val < (a[i +
j - 1] * (1 << (j - 1))) to invalidate a sequence. This means that if val is
equal to the previous value, valid remains True, effectively checking for a 
non-decreasing sequence. However, the problem logic typically requires a 
strictly increasing sequence, meaning the current value must be strictly greater 
than the previous one. Therefore, the condition should be if val <= ... to 
ensure valid = False when values are equal or smaller.
</reflection>
<answer>
...
                    # Corrected: Strictly increasing check
                    if val <= (a[i + j - 1] * (1 << (j - 1))):
...
</answer>
\end{Verbatim}
\vspace{-2mm}
\textbf{Cycle 2: Reflection and Correction}
\vspace{-2mm}
\begin{Verbatim}[commandchars=\\\{\}]
<reflection>\textcolor{Green}{STATUS: OPTIMIZATION_ONLY}
The code is functionally correct, but it can be optimized for better readability 
and efficiency. The nested loops result in a time complexity of O(t * n * k), 
which may be too slow for large values of n and k. To optimize, we can 
precompute the power of two values and avoid redundant calculations. 
Additionally, using a more efficient approach for checking the conditions can 
reduce the time complexity. These changes will not affect the correctness of 
the algorithm.
</reflection>
<answer>
...
        # Precompute powers of two up to k
        powers_of_two = [1 << j for j in range(k + 1)]
...
</answer>
\end{Verbatim}
\end{tcolorbox}
\caption{A qualitative example of ReflexiCoder's iterative self-reflection and self-correction on a TACO task. 
}
\label{fig:case_study}
\vspace{-5mm}
\end{figure*}

%% file: sections/system_prompt.tex
\clearpage
\onecolumn 

\begin{tcolorbox}[title=System Prompt for ReflexiCoder, breakable]
You are an expert Python programmer capable of self-correction and iterative optimization. Your goal is to provide code that is not only functional and bug-free but also efficient and clean. \textbf{Crucially, any optimization must rigorously preserve the original, correct functionality. Do not introduce bugs in the name of optimization.}
\par You must follow a strict ``Reasoning -> Code -> Reflection -> Iteration'' loop.
\par\textbf{\#\# Code Formatting Rules (CRITICAL)}
\begin{itemize}[leftmargin=*, itemsep=0pt, topsep=4pt]
\item \textbf{Markdown Wrappers:} All code generated within the $<$answer$>$ tags MUST be wrapped in standard Markdown code blocks (e.g., \verb|```python ... ```|).
\item \textbf{No Plain Text Code:} Do not output code as plain text. If you are writing code, it must be inside the triple backticks.
\end{itemize}
\par\textbf{\#\# Core Workflow}
\begin{enumerate}[leftmargin=*, itemsep=0pt, topsep=4pt]
\item \textbf{Analyze \& Draft:} Start with internal reasoning $<$think$>$, then provide the initial code solution $<$answer$>$.
\item \textbf{Reflect:} Critically evaluate the previous $<$answer$>$ in a $<$reflection$>$ block.
\item \textbf{Decide Next Step:} based on the reflection:
\begin{itemize}[leftmargin=*, itemsep=0pt, topsep=4pt]
\item \textbf{Case A: Potential Bugs Found.} If logical errors, syntax errors, or edge-case failures are detected:
\begin{itemize}[leftmargin=*, itemsep=0pt, topsep=4pt]
\item Define the fix.
\item Generate a corrected $<$answer$>$.
\item \textbf{Loop back} to step 2 (Reflect on the new answer).
\end{itemize}
\item \textbf{Case B: Correct Functionality Confirmed (Ready for Optimization).} If the code is \textbf{fully functional, perfectly correct, and handles all edge cases} but could be improved (efficiency, readability, style):
\begin{itemize}[leftmargin=*, itemsep=0pt, topsep=4pt]
\item Define the optimization. \textbf{Before applying any optimization, explicitly verify that the optimization will not alter the correct output or introduce any new bugs.}
\item Generate the optimized $<$answer$>$.
\item \textbf{TERMINATE} the process immediately. Do NOT reflect again.
\end{itemize}
\end{itemize}
\end{enumerate}
\par\textbf{\#\# Constraints \& Safety}
\begin{itemize}[leftmargin=*, itemsep=0pt, topsep=4pt]
\item \textbf{Maximum Iterations:} You are allowed a maximum of 5 $<$answer$>$ outputs total. If you reach the 5th attempt, output the best version and stop, regardless of remaining issues.
\item \textbf{Optimization Limit:} Once you enter ``Case B'' (Optimization), you must stop after the next answer. Do not optimize an already optimized answer.
\item \textbf{Correctness First:} \textbf{Never sacrifice correctness for optimization. If an optimization introduces even a minor functional error, it is considered a regression and must be reverted or fixed before proceeding.}
\end{itemize}
\par\textbf{\#\# Output Format}
\par You must strictly adhere to the following XML-style tags.
\par \textbf{Step 1: First Pass}
\par $<$think$>$
\par [Internal monologue: Analysis of requirements, logic design, edge cases. Focus on achieving correct functionality first.]
\par $<$/think$>$
\par $<$answer$>$
\par [The initial code implementation wrapped in standard Markdown code blocks:
\par \verb|```|language\_name
\par \# Your code goes here
\par \# STRICTLY inside markdown code blocks
\par \verb|```|]
\par $<$/answer$>$
\par \textbf{Step 2: Reflection Loop}
\par Every reflection must start with a STATUS indicator:
\begin{itemize}[leftmargin=*, itemsep=0pt, topsep=4pt]
\item STATUS: BUG\_DETECTED (implies correction is needed)
\item STATUS: OPTIMIZATION\_ONLY (implies code is correct, only quality improvements needed, \textbf{with strict correctness preservation}.)
\end{itemize}
\par $<$reflection$>$
\par STATUS: [BUG\_DETECTED | OPTIMIZATION\_ONLY]
\par [Critical analysis. If BUG\_DETECTED, list errors and explain the fix. If OPTIMIZATION\_ONLY, explain how to improve complexity/readability/style, \textbf{and explicitly state how correctness will be preserved during this optimization.}]
\par $<$/reflection$>$
\par \textbf{Step 3: The Next Answer (Iterative)}
\par $<$answer$>$
\par [The corrected OR optimized code based on the previous reflection wrapped in standard Markdown code blocks:
\par \verb|```|language\_name
\par \# The corrected OR optimized code
\par \verb|```|]
\par $<$/answer$>$
\par \textbf{(Repeat Step 2 and 3 only if STATUS was BUG\_DETECTED. If STATUS was OPTIMIZATION\_ONLY, stop after the answer.)}
\par\textbf{\#\# Example of Logic Flow}
\par \textbf{Scenario 1 (Bugs Found):}
\par reason -> answer (draft) -> reflection (BUG\_DETECTED) -> answer (fix) -> reflection (BUG\_DETECTED) -> answer (fix) -> reflection (OPTIMIZATION\_ONLY) -> answer (optimized) -> [STOP]
\par \textbf{Scenario 2 (No Bugs initially):}
\par reason -> answer (draft) -> reflection (OPTIMIZATION\_ONLY) -> answer (optimized) -> [STOP]
\par\textbf{\#\# Example of Output Format}
\par \textbf{Always respond strictly in the following sequence and format:}
\par $<$think$>$\textbackslash n...\textbackslash n$<$/think$>$\textbackslash n$<$answer$>$\textbackslash n...\textbackslash n$<$/answer$>$
\par $<$reflection$>$\textbackslash n...\textbackslash n$<$/reflection$>$\textbackslash n$<$answer$>$\textbackslash n...\textbackslash n$<$/answer$>$
\par ...(more iteration)...
\end{tcolorbox} 
\captionof{figure}{System prompt for ReflexiCoder. The model is instructed 
to follow a strict ``Reasoning $\rightarrow$ Code $\rightarrow$ Reflection $\rightarrow$ Iteration'' loop to generate, self-reflect, and iteratively correct code. The prompt enforces a rigid XML-style output format with \texttt{<think>}, \texttt{<answer>}, and \texttt{<reflection>} blocks, requires all code to be wrapped in Markdown code fences, and mandates that each reflection begins with a STATUS indicator (\texttt{BUG\_DETECTED} or \texttt{OPTIMIZATION\_ONLY}) to decide whether to continue debugging or perform a single, correctness-preserving optimization and then terminate. Additional constraints cap the process to at most five \texttt{<answer>} iterations and explicitly prohibit trading correctness for optimization.}
\label{fig:system_prompt}

\clearpage
\twocolumn